\documentclass[pdflatex,sn-mathphys-num]{sn-jnl}

\usepackage{graphicx}%
\usepackage{multirow}%
\usepackage{amsmath,amssymb,amsfonts}%
\usepackage{amsthm}%
\usepackage{mathrsfs}%
\usepackage[title]{appendix}%
\usepackage{xcolor}%
\usepackage{textcomp}%
\usepackage{manyfoot}%
\usepackage{booktabs}%
\usepackage{algorithm}%
\usepackage{algorithmicx}%
\usepackage{algpseudocode}%
\usepackage{listings}%

\theoremstyle{thmstyleone}%
\theoremstyle{thmstyletwo}%

\theoremstyle{thmstylethree}%

\begin{document}

\title[Prevalence and Binary Classification Model Evaluation]{Area under the ROC Curve has the Most Consistent Evaluation for Binary Classification}


\author*[1]{\fnm{Jing} \sur{Li}}\email{jingl8@illinois.edu}



\affil*[1]{\orgname{University of Illinois at Urbana-Champaign}, \orgaddress{\street{420 David Kinley Hall, 1407 W Gregory Drive}, \city{Urbana}, \postcode{61801}, \state{Illinois}, \country{USA}}}






\abstract{The proper use of model evaluation metrics is important for model evaluation and model selection in binary classification tasks. This study investigates how consistent different metrics are at evaluating models across data of different prevalence while the relationships between different variables and the sample size are kept constant. Analyzing 156 data scenarios, 18 model evaluation metrics and five commonly used machine learning models as well as a naive random guess model, I find that evaluation metrics that are less influenced by prevalence offer more consistent evaluation of individual models and more consistent ranking of a set of models. In particular, Area Under the ROC Curve (AUC) which takes all decision thresholds into account when evaluating models has the smallest variance in evaluating individual models and smallest variance in ranking of a set of models. A close threshold analysis using all possible thresholds for all metrics further supports the hypothesis that considering all decision thresholds helps reduce the variance in model evaluation with respect to prevalence change in data. The results have significant implications for model evaluation and model selection in binary classification tasks.}




\maketitle

\section{Introduction}

The proper use of evaluation metrics for binary classification tasks have received considerable scientific attention in recent years especially for imbalanced data \citep{Powers2011EvaluationFP, Garcia_suitability_2012, lever_classification_2016, luque_impact_2019, jadhav_novel_2020, zhu_performance_2020, chicco_advantages_2020, chicco_matthews_2021, de_diego_general_2022, hicks_evaluation_2022, lavazza_common_2023, chicco2023}. In particular, Area under the Receiver Operating Characteristic (ROC) Curve (AUC) as the most widely used metric in model evaluation has received the most scholarly attention. 

For binary classification, the Receiver Operating Characteristic (ROC) Curve considers all decision thresholds and plots the corresponding false positive rate (FPR) and true positive rate (TPR) on the x-axis and y-axis respectively. The Area under the ROC Curve (AUC) essentially depicts the probability that a randomly selected positive case ranks higher than a randomly selected negative case, and has one-to-one equivalence to the Wilcoxon statistic \citep{hanley_meaning_1982, metz_basic_1978, Zweig1993ReceiveroperatingC, hoo_what_2017}. In comparison with accuracy and other metrics, AUC has shown a number of advantages: increased sensitivity in Analysis of Variance (ANOVA) tests; decreasing standard error with increasing sample size; threshold-independent and invariant to priori class probabilities \citep{bradley_use_1997}. As a result, AUC as a evaluation metric has a dedicated history not only for model evaluation but also for medical screening test such as early cancer detection \citep{baker_central_2003}, bankruptcy prediction \citep{perez-pons_evaluation_2022}, software defect prediction \citep{morasca_assessment_2020}, and many other purposes\citep{parodi_roc_2003, taha_metrics_2015, yang_receiver_2017}, albeit the use of AUC has faced more scrutiny recently\citep{berrar_caveats_2012, cook_use_2007, powers_problem_2012, muschelli_roc_2020}. 

Critics of AUC have pointed out at least two issues associated with its use: it does not account for prevalence and does not work well with imbalanced data where either the positive or the negative class overwhelmingly dominates the AUC score \citep{halligan_disadvantages_2015, wald_is_2014, movahedi_limitations_2023}; not all regions of AUC are meaningful or represent realistic evaluation outcomes \citep{lobo_auc_2008, mcclish_analyzing_1989, lavazza_considerations_2022}. As a result, a number of alternatives have been proposed. For example, Matthew's correlation coefficient (MCC) has been proposed as a better alternative because it has high values only when a model predicts well on both positive and negative classes \citep{chicco_advantages_2020, chicco_matthews_2021, chicco_benefits_2021, chicco2023}.

Additional proposed alternatives to the standard AUC include: average positive predictive value (AP) that varies with respect to prevalence in the data and places more emphasis on early positives than does AUC\citep{Yuan15}; uniform AUC to make AUC not dependent on the range of predicted values \citep{Valverde21}; not proper ROC Curve to identify hidden subclass in microarray data \citep{parodi_not_2008}; restricted ROC Curves to separate informative from noninformative values \citep{parodi_restricted_2016}; concentrated AUC to address the early retrieval problem \citep{swamidass_croc_2010}; partial AUC metrics \citep{vivo_rethinking_2018, carrington_new_2020}; delimited region of interest in the ROC space \citep{mcclish_analyzing_1989, lavazza_considerations_2022}; total operating characteristic \citep{pontius_jr_total_2014}; expected cost representation \citep{Drummond2000}; precision-recall curve to better suit low-prevalence data \citep{ozenne_precisionrecall_2015, movahedi_limitations_2023}; general performance score to combine several user-selected metrics \citep{de_diego_general_2022, Redondo20}, etc. 

In particular, one research \citep{chicco2023} advises "MCC as the standard binary classification metric for any scientific study in any scientific field" because it is the only metric that has high values when true positive rate, true negative rate, positive predictive value and negative predictive value are all high. However, MCC, as the Pearson correlation coefficient of two binary variables, is a quite conservative evaluation metric and requires the two sequences of predicted and actual class labels to highly align with each other in order to have high values. For example, suppose the true class labels of 10 observations are (1, 0, 0, 0, 0, 1, 1, 0, 0, 1), and the predicted class labels are (1, 1, 0, 0, 0, 1, 1, 0, 0, 1). While the classification misses only one observation, the corresponding MCC is 0.82. If the predicted class labels are (0, 0, 0, 0, 0, 1, 1, 0, 0, 0) and the classification misses only two observations, MCC is only 0.61. If the predicted class labels are (0, 0, 1, 0, 0, 0, 1, 0, 0, 1) and the classification misses three observations, MCC is only 0.36 even when there is 70 percent accuracy. For MCC, not only does the number of correct classifications matter, but the order in which the correct and incorrect predictions happen matters significantly.

An existing study \citep{zhu_performance_2020} also investigates the behavior of MCC by letting either true positive rate or true negative rate change while holding the other one constant. The results show that MCC is disproportionately concentrated in the low value range of -0.1 and 0.1 when data is highly class imbalanced even when the overall accuracy is high, an unsurprising result given the conservative nature of MCC. In addition, the study results show that classification accuracy involving positive predictive value and negative predictive value behaves undesirably for imbalanced data because of their formulation. As has been emphasized already \citep{Zweig1993ReceiveroperatingC}, positive predictive value and negative predictive value are not properties of model performance or diagnostic test itself. They are more of an additional interpretation of a given test or classification result rather than a measure of intrinsic model performance. Therefore, a well-performing classification model does not need to always have high positive predictive and negative predictive values. Existing research also shows that AUC is more discriminating than MCC for both balanced and imbalanced data sets \citep{halimu_empirical_2019}.

In general, there are at least three limitations associated with existing research on model evaluation for binary classification and the use of AUC in particular. One is that existing research mostly focuses on model evaluation on one data set and fails to consider the consistency of model evaluation across data sets with varying prevalence levels. A related issue is that existing research does not sufficiently separate out influence of the model being trained and sample size when assessing the proper use of evaluation metrics. Moreover, for metrics other than AUC, a close examination of the effect of decision threshold used and the number of decision thresholds used on evaluation metrics still does not exist.

This paper revisits model evaluation metrics for binary classification, and studies \textbf{to what extent different metrics give consistent evaluation of model performance over data of varying prevalence} while the relationships between different variables and sample size are kept constant. We measure consistency of model evaluation using variance of metric values for individual models and variance of rankings for a set of models across data of different prevalence. Using statistical simulation, we conduct analysis for 18 evaluation metrics and five commonly used machine learning models as well as a naive random guess model. The results show that evaluation metrics that are less influenced by prevalence offer more consistent evaluation of individual models and more consistent ranking of a set of models. In particular, AUC has the smallest variance with respect to prevalence and the smallest variance in its ranking of different models. We further hypothesize that the reason AUC is able to achieve more consistent evaluation is because it considers all decision thresholds, and a threshold analysis considering all possible thresholds support this claim. 

Consistency of model evaluation across data of different prevalence is important because repeated evaluations of same model across data of different prevalene are common. And we don't want model evaluation results change simply because prevalence in the data changes and the metric used has a close to monotonic relationship with prevalence. If that is the case, evaluation metric values become artifact of the metrics being used rather than measures of intrinsic model performance. As a result, practitioners of model evaluation are left unsure about the true predictive power of the model being used. This can have undesirable consequences for model evaluation and model selection in various domains \citep{yao_assessing_2020}. Therefore, it is essential that model evaluation results are consistent with prevalence change in the data.

The novelty of this paper resides in its focus on an underappreciated question in model evaluation and its use of statistical simulation with a real-world data set, which generates a much richer set of data scenarios compared with existing research (having only a small number of cases). Thus, we show a more complete picture of the relationships between data, evaluation metric and model while keeping sample size and the relationships between variables constant. In addition, a much larger set of evaluation metrics are considered  here than existing research and an explicit link is made between prevalence, evaluation of individual models, and ranking of a set of models by different metrics. Moreover, we perform an extensive analysis of the effect of decision threshold on model evaluation outcomes and show that using a larger number of thresholds can improve model evaluation consistency.

This paper is organized as follows, in section 2, we first provide background information and notations on model evaluation metrics for binary classification. In section 3, we look at a prominent case of predicting crime recidivism and present a simulation study that examines how different confusion matrix metrics evaluate models differently under data with different prevalence. In section 4, we show the analysis results. And in section 5, we offer discussions and conclusions.

\section{Methods}

\subsection{Background and Basic Notations}\label{sec2.1}

Evaluation metrics are statistics used to evaluate model performance especially for out of sample prediction tasks as is commonly the case for machine learning model applications. Below we compare several evaluation metrics and look at how accurate and reliable they are at capturing the model prediction performance for data of different prevalence level. We focus on binary classification tasks in this study. First, let us introduce a few concepts. For every binary prediction/classification task, the result could fall into four different categories:
\begin{itemize}
\item true positives (TP), positive cases correctly predicted as positive;
\item false negatives (FN), positive cases incorrectly predicted as negative;
\item true negatives (TN), negative cases correctly predicted as negative;
\item false positives (FP), negative cases incorrectly predicted as positive.
\end{itemize}

The partition of these four categories can be represented in a $2 \times 2$ tabular format called confusion matrix: $\begin{pmatrix}
  TP & FN\\ 
  FP & TN
\end{pmatrix}$, TP + FN + FP + TN = n. 

The first set of prediction performance metric we look at are: 
\begin{equation*}
    \begin{aligned}
\text{accuracy} = \frac{\text{TP} + \text{TN}}{\text{TP} + \text{FN} + \text{TN} + \text{FP}} \\ \text{True Positive Rate (TPR)} = \frac{\text{TP}}{\text{TP} + \text{FN}} \\ \text{True Negative Rate (TNR)} = \frac{\text{TN}}{\text{TN} + \text{FP}} \\ \text{Positive predictive value (PPV)} = \frac{\text{TP}}{\text{TP} + \text{FP}} \\ \text{Negative predictive value (NPV)} = \frac{\text{TN}}{\text{TN} + \text{FN}}
    \end{aligned}
\end{equation*}

And a number of metrics that combine TPR and TNR in certain ways: 
\begin{equation*}
    \begin{aligned}
\text{Balanced Accuracy (BA)} = \frac{\text{TPR} + \text{TNR}}{2} \\ \text{Bookmaker informedness (BI)} = \text{TPR} + \text{TNR} - 1 \\ \text{Markedness (MK)} = \text{PPV} + \text{NPV} - 1 \\ \text{geometric mean (Gmean)} = \sqrt{\text{TPR} \times \text{TNR}} \\ \text{Jaccard index (JI)} = \frac{\text{TP}}{\text{TP} + \text{FN} + \text{FP}} \\ \text{diagnostic odds ratio (DOR)} = \frac{\text{TPR} \times \text{TNR}}{ \text{FPR} \times \text{FNR}} \\ \text{Fowlkes-Mallows index (FM)} = \sqrt{\text{PPV} \times \text{TPR}} \\ 
\text{Cohen's Kappa} = \frac{\text{Accuracy} - \text{expAccuracy}}{1 - \text{expAccuracy}}, \\ \text{expAccuracy} = \frac{(\text{TP} + \text{FN}) \cdot (\text{TP} + \text{FP}) + (\text{TN} + \text{FP}) \cdot (\text{TN} + \text{FN}) }{n^2}
    \end{aligned}
\end{equation*}.

F-scores\footnote{In the following we use $\beta = 2, 0.5$, two commonly used values.} and Matthew's correlation coefficient (MCC): 
\begin{equation*}
\begin{aligned}
\text{F1 score} = \frac{2}{\text{recall}^{-1} + \text{precision}^{-1}} =  \frac{2TP}{2TP + FP + FN} \\ F_{\beta}~\text{score} = (1 + \beta^2) \cdot \frac{\text{PPV} \cdot \text{TPR}}{(\beta^2 \cdot \text{PPV}) + \text{TPR}} \\
\text{MCC} = \frac{\text{TP} \cdot \text{TN} - \text{FP} \cdot \text{FN}}{\sqrt{(\text{TP} + \text{FP})\cdot(\text{TP} + \text{FN})\cdot(\text{TN} + \text{FP})\cdot(\text{TN} + \text{FN})}} \\ = \scriptstyle{\sqrt {\text{PPV} \times \text{TPR} \times \text{TNR} \times \text{NPV}} - \sqrt{\text{FDR} \times \text{FNR} \times \text{FPR} \times \text{FOR}}} ~\text{(minimum: -1, maximum: +1)}, \\ \text{false omission rate (FOR)} = 1 - \text{NPV} \\ \text{false discovery rate (FDR)} = 1 - \text{PPV} \\ \text{False Positive Rate (FPR)} = 1 - \text{TNR} \\ \text{False Negative Rate (FNR)} = 1 - \text{TPR}
\end{aligned}
\end{equation*}. 

As will become clearer in subsequent analysis, each of these metrics emphasizes different aspects of the performance of a binary classifier. For example, metrics such as the F-scores, weigh heavily on TPR and will be most informative if the particular classification task cares most of the ability to detect true positive cases while the opposite is true for metrics that weigh heavily on TNR. A number of metrics including balanced accuracy, bookmaker informedness, geometric mean and MCC, in their formulations, try to balance model performance on both classes of data and are most useful when the purpose of model evaluation is to assess accuracy on both classes of data equally.

In addition, some metrics originate from specific application domains and can have their particular use cases. In particular, bookmaker informedness in the binary case measures how much a test or model is better than random guessing and is primarily used in medical diagnostic test. Markedness is a measure that comes from linguistics, and measures the extent to which a unit is different from another unit as abnormal or divergent. Jaccard index, also called intersection over union, is a measure of similarity or diversity between two sets of values, and a common metric for image object recognition in computer vision. Diagnostic odds ratio measures the effectiveness of medical diagnostic test, is the ratio between the odds for a test result being positive if the unit has disease and the odds for a test result being positive if the unit has no disease. Fowlkes-Malloows Index is the square root of the product of TPR and PPV, and measures the similarity between two clustering/classification outcomes. Cohen's Kappa, a measure of inter-rater reliability from psychology, measures the agreement between two raters each classifying $N$ items into $C$ categories. F1 score is the harmonic mean of TPR and PPV while the $F_\beta$ scores treat TPR as $\beta$ times more important than PPV. The F-scores are commonly used in information retrieval and named entity recognition and word segmentation tasks in natural language processing.

\subsection{Alternative Derivation of Confusion Matrix Metrics}\label{sec2.2}

As mentioned in the literature \citep{Kruschke_2015, luque_impact_2019}, all entries of a confusion matrix and confusion matrix metrics can be derived from the four quantities of sample size $n$, prevalence, $\text{TPR}$ and $\text{TNR}$. To show that this is indeed the case, below we redefine the confusion matrix metrics introduced above in terms of these four quantities excluding metrics that are already expressed in these four quantities. We use $\phi$ to represent prevalence, and $\phi = \frac{TP + FN}{n}$.

\begin{equation}\label{eq:1}
\begin{aligned}
\text{accuracy} = \frac{\text{TP} + \text{TN}}{n} = \text{TPR} \cdot \phi + \text{TNR} \cdot (1 - \phi)
\end{aligned}
\end{equation}

\begin{equation}\label{eq:2}
\begin{aligned}
\text{F1 score} = \frac{2TP}{2TP + FP + FN} = \frac{2\text{TPR} \cdot \phi \cdot n}{2\text{TPR} \cdot \phi \cdot n + (1 - \text{TNR}) \cdot (1 - \phi) \cdot n + (1 - \text{TPR}) \cdot \phi \cdot n} \\ = \frac{2} {2  + \frac{1 - \text{TNR}}{\text{TPR}} \frac {1 - \phi}{\phi}  + \frac {1 - \text{TPR}}{\text{TPR}}} = \frac{2} {2  + \frac{1}{\text{TPR}}((1 - \text{TNR}) \frac {1 - \phi}{\phi}  + 1 - \text{TPR})}
\end{aligned}
\end{equation} 

\begin{equation}\label{eq:3}
\begin{aligned}
\text{PPV} = \frac{\text{TPR} \cdot \phi \cdot n}{\text{TPR} \cdot \phi \cdot n + (1 - \text{TNR}) \cdot (1 - \phi) \cdot n} = \frac{\text{TPR} \cdot \phi }{\text{TPR} \cdot \phi  + (1 - \text{TNR}) \cdot (1 - \phi) }.
\end{aligned}
\end{equation}

\begin{equation}\label{eq:4}
\begin{aligned}
\text{NPV} = \frac{\text{TNR} \cdot (1 - \phi) \cdot n}{\text{TNR} \cdot (1 - \phi) \cdot n + (1 - \text{TPR}) \cdot \phi \cdot n} = \frac{\text{TNR} \cdot (1 - \phi) }{\text{TNR} \cdot (1 - \phi)  + (1 - \text{TPR}) \cdot \phi }.
\end{aligned}
\end{equation}


\begin{equation}\label{eq:5}
\begin{aligned}
\text{Markedness (MK)} = \frac{\text{TPR} \cdot \phi }{\text{TPR} \cdot \phi  + (1 - \text{TNR}) \cdot (1 - \phi) } +  \frac{\text{TNR} \cdot (1 - \phi) }{\text{TNR} \cdot (1 - \phi)  + (1 - \text{TPR}) \cdot \phi } - 1
\end{aligned}
\end{equation}

\begin{equation}\label{eq:6}
\begin{aligned}
\text{Jaccard index (JI)} = \frac{\text{TPR} \cdot \phi }{\phi + (1 - \text{TNR}) \cdot (1 - \phi) }
\end{aligned}
\end{equation}

\begin{equation}\label{eq:7}
\begin{aligned}
\text{diagnostic odds ratio (DOR)} = \frac{\text{TPR} \times \text{TNR}}{ (1 - \text{TNR}) \times (1 - \text{TPR})}        
\end{aligned}
\end{equation}

\begin{equation}\label{eq:8}
    \begin{aligned}
\text{Fowlkes-Mallows index (FM)} = \frac{\text{TPR}^2 \cdot \phi }{\text{TPR} \cdot \phi  + (1 - \text{TNR}) \cdot (1 - \phi) }
    \end{aligned}
\end{equation}

\begin{equation}\label{eq:9}
    \begin{aligned}
\text{Kappa} = \frac{\text{TPR} \cdot \phi + \text{TNR} \cdot (1 - \phi) - \text{expAccuracy}}{1 - \text{expAccuracy}}, \\ \text{expAccuracy} = \frac{\phi^2\text{TPR} +  \phi (1 - \phi)(2 - \text{TPR} - \text{TNR}) + (1 - \phi)^2\text{TNR}}{n}
    \end{aligned}
\end{equation}

\begin{flalign}\label{eq:10}
\text{MCC} &= \scriptstyle{\sqrt { \frac{\text{TPR}^2 \cdot \phi }{\text{TPR} \cdot \phi  + (1 - \text{TNR}) \cdot (1 - \phi) } \cdot \frac{\text{TNR}^2 \cdot (1 - \phi) }{\text{TNR} \cdot (1 - \phi)  + (1 - \text{TPR}) \cdot \phi }}}  - \scriptstyle{\sqrt{\frac{(1 - \text{TNR})^2 \cdot (1 - \phi) }{\text{TPR} \cdot \phi  + (1 - \text{TNR}) \cdot (1 - \phi) } \cdot \frac{(1 - \text{TPR})^2 \cdot \phi}{\text{TNR} \cdot (1 - \phi)  + (1 - \text{TPR}) \cdot \phi }}} && \\\nonumber
&= \frac{\text{TPR} + \text{TNR} - 1}{\sqrt{(\text{TPR} \cdot \frac{\phi}{1 - \phi}  + 1 - \text{TNR})(\text{TNR} \cdot \frac{1 - \phi}{\phi} + 1 - \text{TPR})}} &&.
\end{flalign}

\section{Case Study: Crime Recidivism}

First, we look at a real-world case where the outcome is more or less class balanced. We leverage the Broward County data set \citep{dressel_accuracy_2018, bansak_can_2019} that consists of individuals arrested in Broward County, Florida between 2013 and 2014. The sample size is 6214, with 2775 positive cases, individuals who reoffended and 3439 negatives, individuals who did not reoffend. The predictors for the classifier include seven features of the defendants: gender, age, number of juvenile misdemeanors, number of juvenile felonies, number of prior (nonjuvenile) crimes, crime degree, and crime charge \citep{dressel_accuracy_2018}. We train five machine learning models and see how the different evaluation metrics listed above capture the prediction performance of these models: logistic regression (GLM), random forest (RF), k-nearest neighbors (KNN), linear discriminant analysis (LDA), gradient boosting machine (GBM). In addition, we make a random guess model that randomly assign class labels to observations based on prevalence of the test set. For example, for a test set with 60 percent positive cases, random guess will randomly assign the positive label to 60 percent cases and negative label to the other 40 percent\footnote{The probability of positive class is a random number between 0.51 and 0.99 while probability of negative class is a random number between 0.01 and 0.49.}. The sample data are randomly split into 80 percent training data and 20 percent test data. 10-fold cross-validation is used with the model training process \footnote{In short, 10-fold cross-validation splits the training data into 10 random samples, iteratively uses 9 samples to estimate a model and the remaining one sample to validate the model, and choose the model that has the best performance across the 10 random samples. This procedure helps to address over-fitting and find the model that has best predictive power.}. Table ~\ref{table2} lists the rankings of the different models by the different metrics. 

\begin{table}[!htbp]
\centering
\caption{Crime Recidivism Prediction: Rank of Different Models} 
\label{table2}
\begin{tabular}{l|r|r|r|r|r|r}
\hline
Metric & GBM & GLM & KNN & LDA & Random Forest & randomguess\\
\hline
True\_positives & 1 & 4 & 3 & 5 & 2 & 6\\
\hline
False\_negatives & 1 & 4 & 3 & 5 & 2 & 6\\
\hline
True\_negatives & 5 & 3 & 4 & 1.5 & 1.5 & 6\\
\hline
False\_positives & 5 & 3 & 4 & 1.5 & 1.5 & 6\\
\hline
TPR & 1 & 4 & 3 & 5 & 2 & 6\\
\hline
TNR & 5 & 3 & 4 & 1.5 & 1.5 & 6\\
\hline
PPV & 3 & 4 & 5 & 2 & 1 & 6\\
\hline
NPV & 1 & 3 & 5 & 4 & 2 & 6\\
\hline
Accuracy & 1 & 3.5 & 5 & 3.5 & 2 & 6\\
\hline
BA & 1 & 3 & 5 & 4 & 2 & 6\\
\hline
BI & 1 & 3 & 5 & 4 & 2 & 6\\
\hline
F1\_score & 1 & 3 & 4 & 5 & 2 & 6\\
\hline
MCC & 1 & 4 & 5 & 3 & 2 & 6\\
\hline
Gmean & 1 & 3 & 5 & 4 & 2 & 6\\
\hline
Fowlkes\_Mallows & 1 & 3 & 5 & 4 & 2 & 6\\
\hline
Markedness & 2 & 4 & 5 & 3 & 1 & 6\\
\hline
Diag\_odds\_ratio & 2 & 4 & 5 & 3 & 1 & 6\\
\hline
Jaccardindex & 1 & 3 & 4 & 5 & 2 & 6\\
\hline
Cohens\_kappa & 1 & 3 & 5 & 4 & 2 & 6\\
\hline
F\_beta\_score\_0.5 & 1 & 3 & 4 & 5 & 2 & 6\\
\hline
F\_beta\_score\_2 & 1 & 4 & 3 & 5 & 2 & 6\\
\hline
AUC & 1 & 3 & 5 & 4 & 2 & 6\\
\hline
\end{tabular}
\end{table}

Table~\ref{table3} in Appendix lists the corresponding confusion matrix entries for the model predictions (first four rows) and values of the 18 evaluation metrics. The results indicate that all evaluation metrics rank naive random guess as the worst performing model. GBM is most frequently ranked as the best model, random forest is most frequently ranked as the second best. GLM is most frequently ranked as either third or fourth, KNN is most frequently ranked as fifth but occasionally as third or fourth. LDA is ranked between third, fourth, and fifth or 1.5. Ties in ranking are broken by averages. As TP, FN, TN, FP, TPR, TNR concern positive or negative class only for either actual or predicted labels, it is no surprise that rankings are reversed between metrics focusing on TPR only and those focusing on TNR only. AUC calculations are based on all possible thresholds while all other metric calculations are based on the 0.5 cutoff. These patterns can also be observed from the corresponding metric values in Table ~\ref{table3}. 

While this example shows that different metrics can converge in their evaluation of classification outcomes, this is just one single case. And more importantly, this is a case where the positive and negative classes in the data are more or less balanced. To be precise, the corresponding prevalence is 0.452. When data becomes more imbalanced and either the positive or negative class dominates the data, different metrics' evaluation behavior can be very different. To obtain a more complete picture of the dynamics between prevalence level and model evaluation by different metrics, we use statistical simulation to create a series of data sets with varying prevalence. The results below show how prevalence change affects TPR and TNR, which in turn affect different metrics. The sample size $n$ is kept constant such that we are able to separate out the influence of sample size.

\subsection{Simulation Study}\label{sec4.1}

The following simulation is done with the Broward County data set \citep{dressel_accuracy_2018, bansak_can_2019} used above. The original data set has 2775 positive cases and 3439 negative cases, which is equal to a prevalence of 0.452 \footnote{Here, prevalence refers to prevalence of the sample data as a whole. However, all the above definitions of evaluation metrics in \ref{sec2.1} refer to prevalence of the test set. In this study, as the test sets are randomly sampled from the sample data,  prevalence of the test sets are the same as the sample data (differences are negligible). Prevalence of the simulation results presented below is for the test sets.}. To simulate the dynamics associated with changing prevalence, we apply up-sampling and down-sampling at the same time but for different classes of the data. First, we iteratively drop 30 randomly sampled positive cases while simultaneously adding 30 randomly sampled negative cases, thus incrementally reducing prevalence while keeping the sample size $n$ constant. In total, we run 76 such iterations, which along with the original data sample give us 77 data sets. The 77th data sample has 495 positive cases and 5719 negative cases, equaling a prevalence of about 0.08. We stop at this iteration because TPR decreases to almost 0 while TNR increases to almost 1. Next, we do the opposite. From the original data sample, we iteratively drop 30 randomly sampled negative cases while simultaneously adding 30 randomly sampled positive cases, thus incrementally increasing prevalence while keeping the sample size $n$ constant. In this second phase, we run 79 iterations and stop when TPR increases to almost 1 while TNR decreases to almost 0. The 79th data sample has 5145 positive cases and 1069 negative cases, equaling prevalence of about 0.83.

As units are randomly dropped from or added to the positive and negative classes, the simulation steps above are able to keep the relationships among the predictors and between the predictors and the outcome largely unchanged. The bivariate correlations among all 8 variables have only minimal changes. The correlation heat maps shown in the Supplementary Information showcase the bivariate correlations among all the variables for the original data as well as 79 simulated data samples (every other ones selected from all data samples). It can be observed that the correlations among all variables largely stay the same across all the simulated data samples (The other 76 simulated data samples have been verified to show the same pattern).

\section{Results}
\subsection{Simulation Results: Evaluation Metrics}\label{sec4.2}

\begin{figure*}[!htbp]
\caption{Model Evaluation by different Metrics at different prevalence}\label{fig:foobar1}
    \includegraphics[width=.49\textwidth]{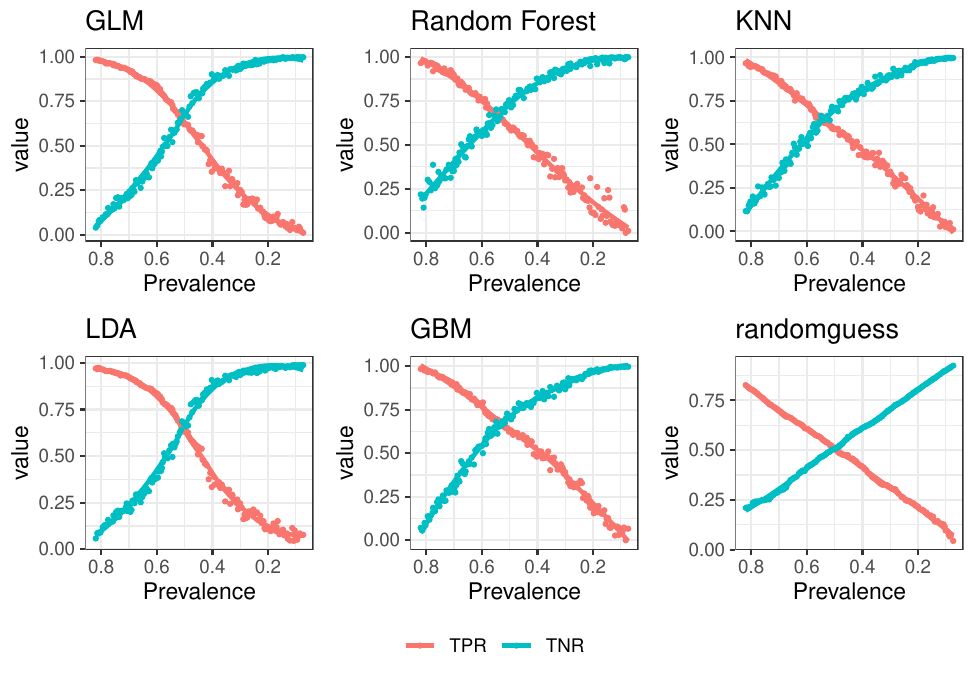}\hfill
    \includegraphics[width=.49\textwidth]{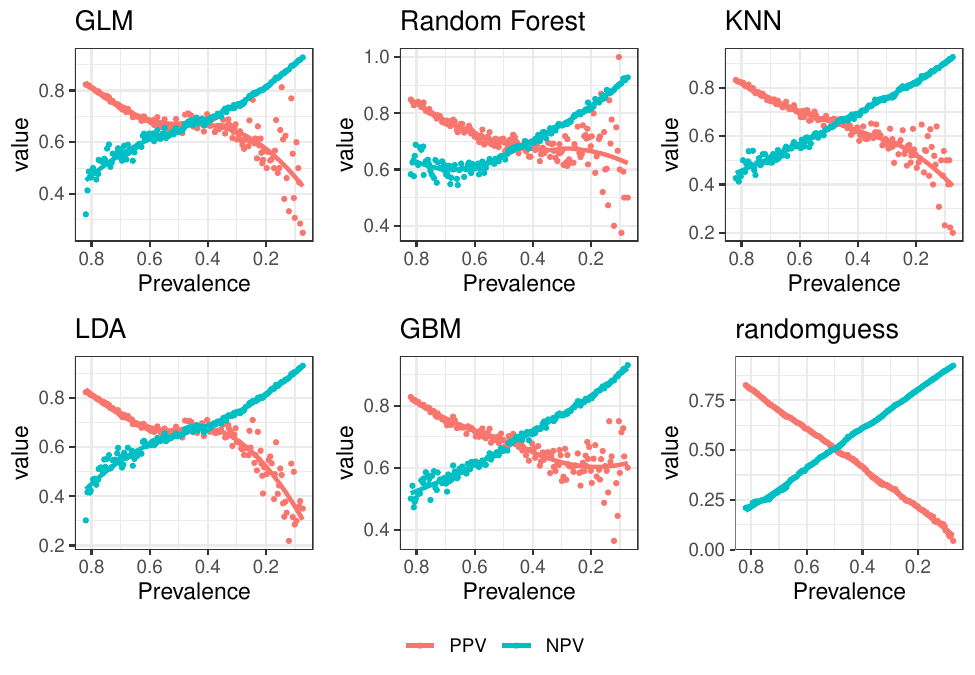}
    \\[\smallskipamount]
    \includegraphics[width=.49\textwidth]{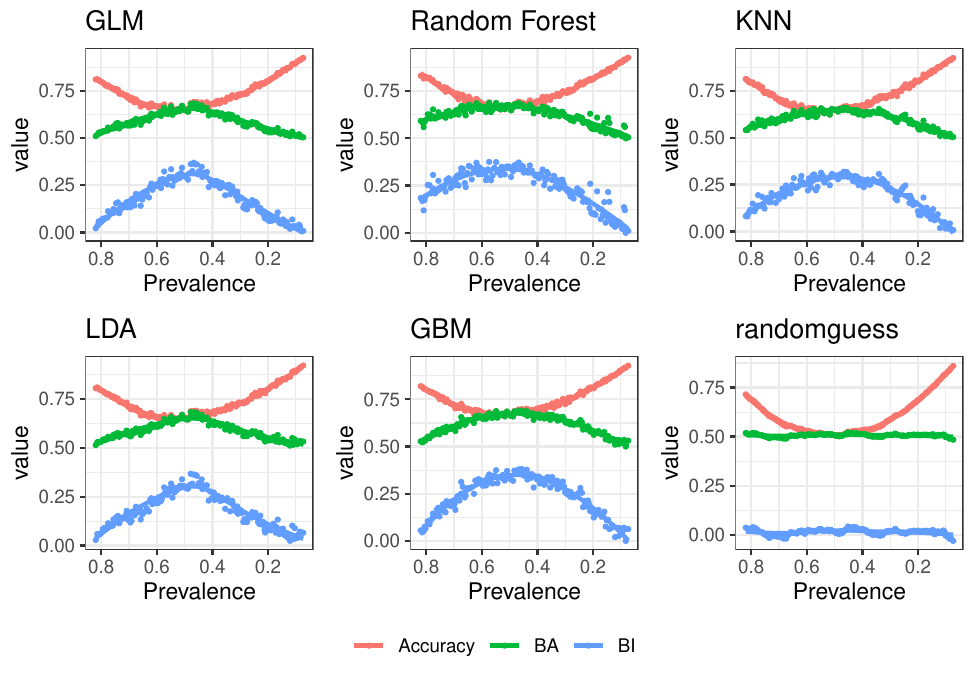}\hfill
    \includegraphics[width=.49\textwidth]{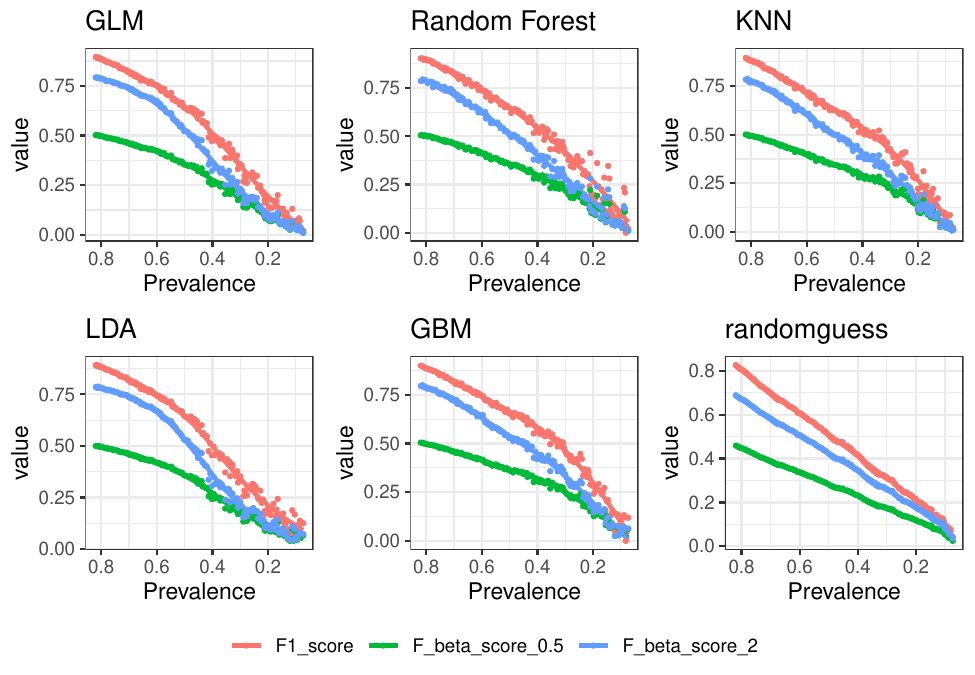}
    \\[\smallskipamount]
    \includegraphics[width=.49\textwidth]{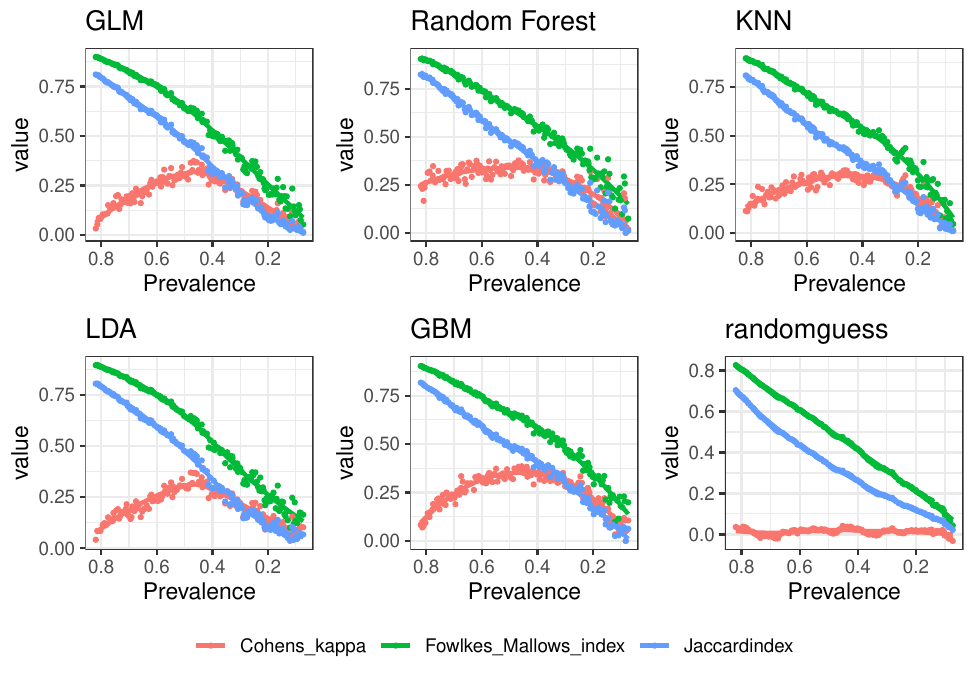}\hfill
    \includegraphics[width=.49\textwidth]{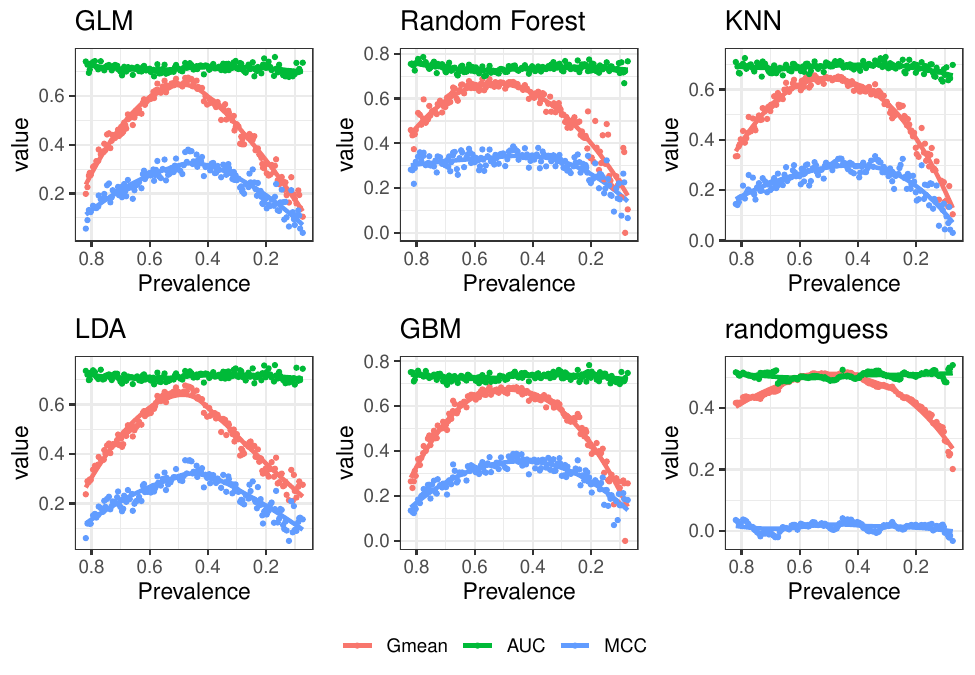}
    \\[\smallskipamount]
    \includegraphics[width=.49\textwidth]{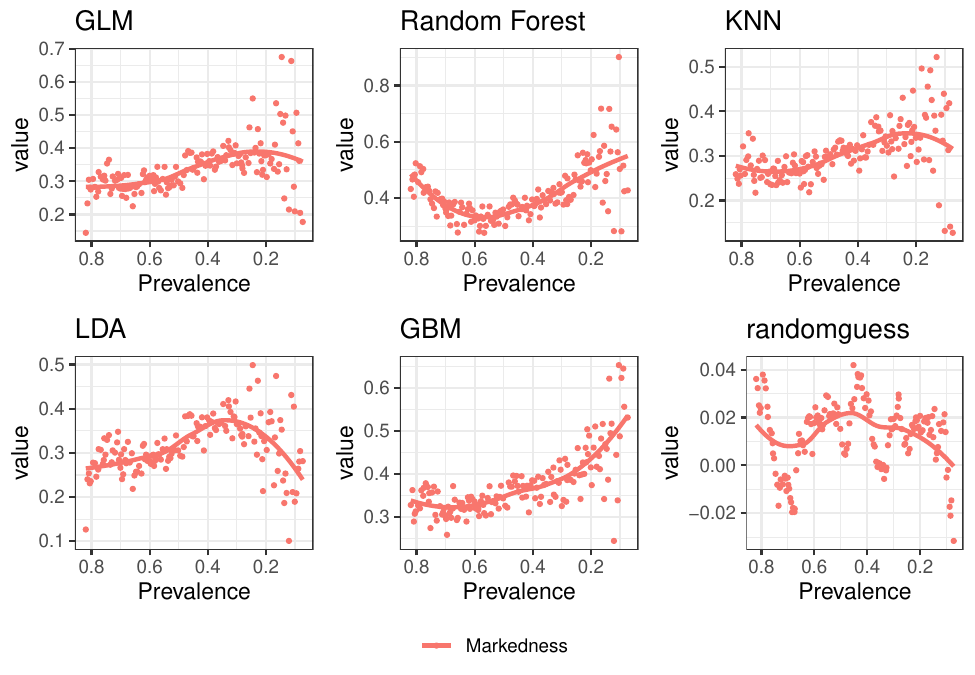}\hfill
    \includegraphics[width=.49\textwidth]{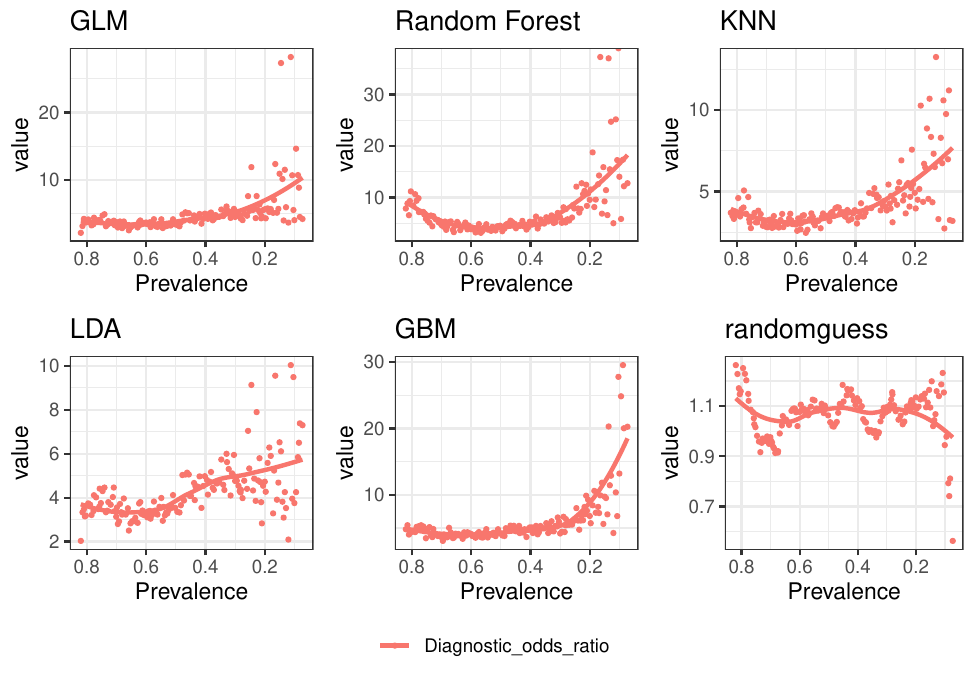}
\end{figure*}

Figure ~\ref{fig:foobar1} shows the relationships between changing prevalence and various model evaluation metrics. The curves are LOESS-smoothed ones. 

First, we look at TPR, TNR, PPV and NPV. Across the six models, decreasing prevalence corresponds with decreasing TPR and increasing TNR. And the two rates intersect at around prevalence 0.5. Similarly, in general, PPV decreases and NPV increases as prevalence decreases. However, for all models except random guess, there are significant fluctuations in PPV when prevalence is low, while NPV shows most fluctuation when prevalence is high. The two curves for PPV and NPV again intersect at around prevalence 0.5, indicating that 0.5 is a critical point for PPV and NPV just as well as TPR and TNR.

Mathematically, from expressions shown in section ~\ref{sec2.2}, we know that $\text{PPV} = \frac{\text{TPR} \cdot \phi }{\text{TPR} \cdot \phi  + (1 - \text{TNR}) \cdot (1 - \phi) } = \frac{1}{1 + \frac{1 - \text{TNR}}{\text{TPR}} \cdot \frac{ 1 - \phi } {\phi}}$. As rate of change in $\text{TPR}$ is similar to the rate of change in $\text{TNR}$ and the rate of change in prevalence across the whole range of prevalence, and as prevalence decreases, $(1 - \text{TNR}) ~ \text{increases}$, $\frac{1}{\text{TPR}} ~ \text{increases}$, $\frac{ 1 - \phi } {\phi} ~ \text{increases}$, therefore, we know that $1 + \frac {1 - \text{TNR}}{\text{TPR}} \cdot \frac{ 1 - \phi } {\phi}$  increases, and $ \text{PPV} ~ \text{decreases}$. 

Similarly, $\text{NPV} = \frac{\text{TNR} \cdot (1 - \phi) }{\text{TNR} \cdot (1 - \phi)  + (1 - \text{TPR}) \cdot \phi } = \frac{1}{1 + \frac{1 - \text{TPR}}{\text{TNR}} \cdot \frac{ \phi}{1 - \phi} } $, and as prevalence decreases, $(1 - \text{TPR}) \text{increases}, \frac{1}{\text{TNR}} \text{decreases}, \frac{\phi}{1 - \phi} \text{decreases} $, and $ \frac{1 - \text{TPR}}{\text{TNR}} \frac{ \phi}{1 - \phi} \text{decreases} $, and $\text{NPV} \text{increases}$. The observed patterns of PPV and NPV in Fig ~\ref{fig:foobar1} largely correspond with their mathematically expected behavior \footnote{Noticeably for changing prevalence, TPR and TNR are more stable than PPV and NPV especially at extreme prevalence. From corresponding expressions in section ~\ref{sec2.1}, we know that denominators of both TPR and TNR are known before predictions are made, therefore, only numerators of TPR and TNR are changing as prevalence changes. For PPV and NPV, both denominator and numerator are only known after predictions are made. Therefore,  for PPV and NPV, both denominator and numerator are changing as prevalence changes. In other words, PPV and NPV vary more because of a particular prediction task while TPR and TNR vary less and are more affected by the independent influence of prevalence, a key distinction not often emphasized in existing literature.}

The three F-scores show a similar downward trend as prevalence decreases. F1 score shows the highest steepness, $F_{\beta}$ score with $\beta$ = 2 next, and $F_{\beta}$ score with $\beta$ = 0.5 shows the least steepness. Apparently, F-scores have a monotonically increasing relationship with prevalence, which is verified by a simple mathematical derivation for F1 score shown in Appendix ~\ref{appendix3}. 

Similarly, both Falkes-Mallows index and Jaccard index show a monotonically increasing relationship with prevalence. From the corresponding expressions of these two metrics in section ~\ref{sec2.2}, we know that both metrics are close to $\text{PPV}$ in form, therefore, it is unsurprising that both metrics show a similarly downtown trend as $\text{PPV}$ though with more steepness. 

It can be observed that accuracy has a convex curve with respect to prevalence, that is accuracy has lower values when prevalence is around 0.5 while it has higher values when prevalence goes to either the upper or lower extremes. Above prevalence 0.5, accuracy is dominated by the positive class and decreasing TPR while below 0.5, accuracy is dominated by the negative class and increasing TNR. Therefore, accuracy is dominated by the majority class. Even random guess shows the same behavior.

On the contrary, both balanced accuracy and bookmaker informedness show a concave curve with respect to prevalence. That is both metrics have higher values when class is more balanced and lower values when class is very imbalanced. This can be understood from their expression as certain form of summation of TPR and TNR. Naive random guess is constant around 0.5 and 0, median of the range for balanced accuracy and bookmaker informedness respectively, indicating the amount of prediction accuracy by random chance. 

Cohen's Kappa shows an concave curve with respect to prevalence. That is it is higher when class is more balanced and lower when class is more imbalanced. From equation \ref{eq:9}, we know that $\text{Kappa} = \frac{\frac{\text{Accuracy}}{\text{expAccuracy}} - 1}{\frac{1}{\text{expAccuarcy}} - 1}$, and $\text{expAccuracy}$ is dominated by decreasing $\text{TPR}$ when prevalence is above 0.5 and by increasing $\text{TNR}$ when prevalence is below 0.5, thus leading to the observed pattern for $\text{Kappa}$. Random guess is constant around 0, median of the range for Kappa indicating the amount of prediction accuracy by random chance.

Similarly, the geometric mean shows a concave curve with respect to prevalence, which can be understood from its expression $\sqrt{\text{TPR} \times \text{TNR} }$. That is it values high TPR and high TNR at the same time. The same applies to MCC, which is essentially the Pearson's correlation coefficient for two binary variables. And to have high values, $\text{MCC}$ requires the two sequence of actual and predicted class labels to closely align with each other. From $\text{MCC} = \sqrt {\text{PPV} \times \text{TPR} \times \text{TNR} \times \text{NPV}} - \sqrt{\text{FDR} \times \text{FNR} \times \text{FPR} \times \text{FOR}}$, we know that $\text{MCC}$ also tries to achieve high $\text{TPR}$ and high $\text{TNR}$ at the same time. Random guess again is constant around 0, median of the range for MCC.

As Area under the ROC Curve (AUC) assesses performance of a classifier at all possible thresholds, it cannot be constructed from a single confusion matrix based on one classification threshold and cannot have an exact comparison with other confusion matrix evaluation metrics calculated with the 0.5 probability cutoff. However, AUC shows a most obvious pattern in Fig ~\ref{fig:foobar1}. AUC is mostly flat regardless of prevalence change except for minor fluctuations at extremely low prevalence.  

Markedness does not show an obvious pattern with prevalence across the six models. As it is a form of sum of PPV and NPV, and both PPV and NPV  fluctuate a lot at extreme prevalence, it is unsurprising to see the large fluctuation in its trends including for random guess. DOR shows a somewhat upward trend at low prevalence. However, for most prevalence levels, it has a flat trend, and there is a lot of fluctuations at low prevalence. This can be understood from its alternative expression $\frac{1}{(1 - \frac{1}{\text{TPR}}) (1 - \frac{1}{\text{TNR}})}$, where the diverging trends of $\text{TPR}$ and $\text{TNR}$ compete with each other.

Here is a summary of the above results. As TPR, TNR, PPV and NPV mostly concern prediction accuracy for one class only, they have a close to monotonic relationship with prevalence in either the positive or negative direction though PPV and NPV fluctuate much more. And a number of metrics appear to have a monotonically increasing relationship with prevalence including the three F-scores, Fowlkes-Mallows index and Jaccardindex. Meanwhile, balanced accuracy, bookmaker informedness, Cohen's Kappa, Matthew's correlation coefficient and geometric mean all have a concave curve with respect to decreasing prevalence though geometric mean has a much larger range of values. Accuracy has a convex curve with respect to decreasing prevalence. AUC is prevalence-independent and shows flat trends across the different models. Markedness and DOR do not have strong patterns in regard to prevalence.

\subsection{Simulation Results: Comparisons of Models}

\begin{figure}[!htbp]
    \caption{\textbf{Model Comparison by different metrics}}\label{fig:foobar3}
    \includegraphics[scale=0.745]{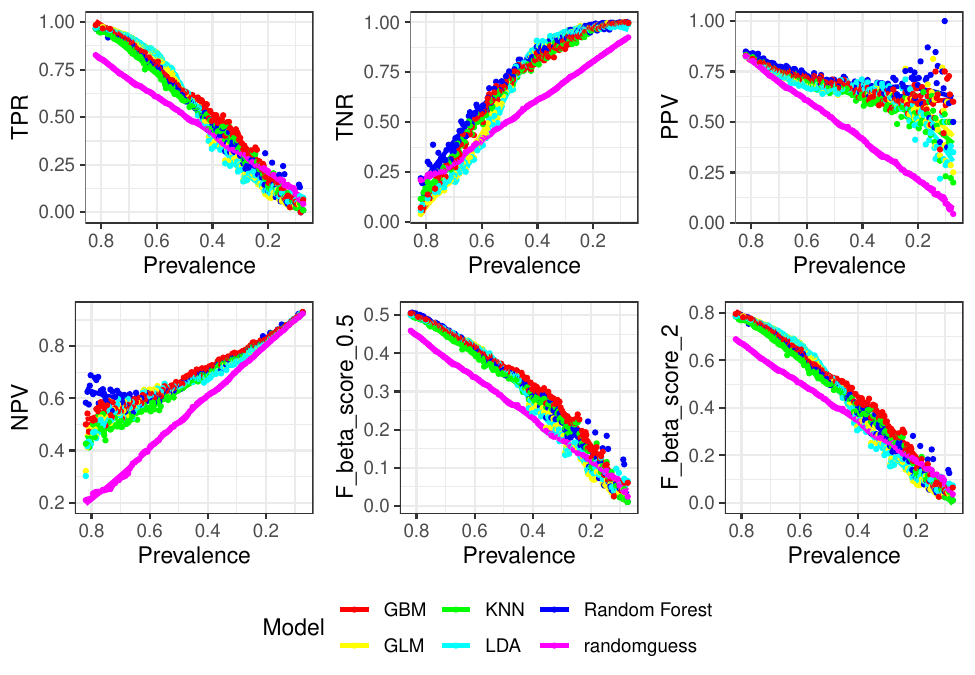}
    \includegraphics[scale=0.745]{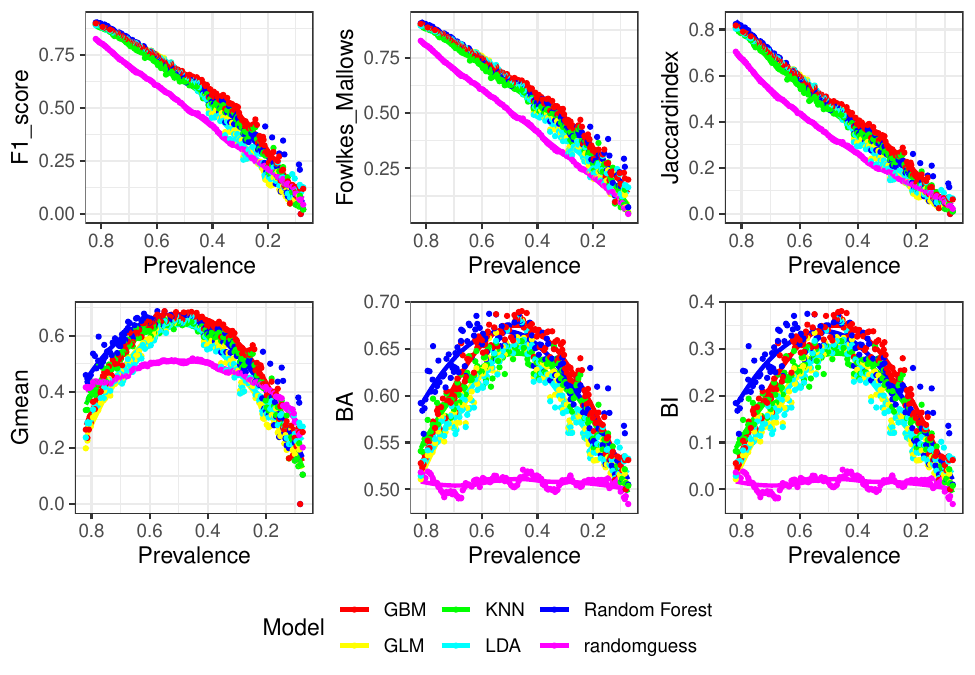}
    \includegraphics[scale=0.745]{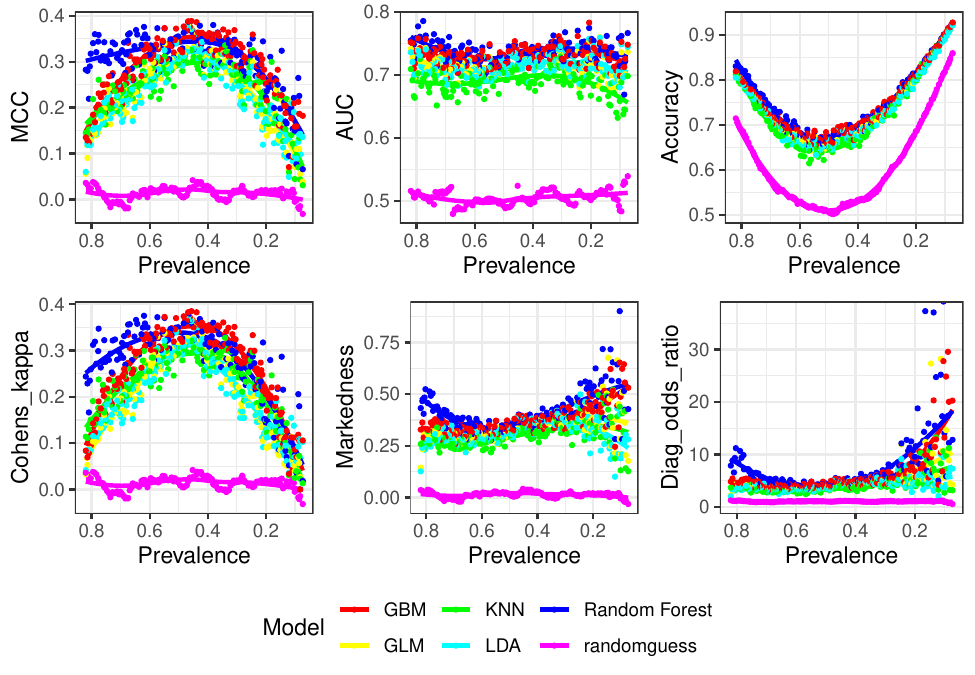}
\end{figure}

Figure ~\ref{fig:foobar3} shows how the 6 models are evaluated by such metrics. For presentation convenience, we organize the 18 metrics into the upper, middle and lower panes. The results indicate that evaluation metrics that are more influenced by prevalence as shown in Fig ~\ref{fig:foobar1} also do a poorer job differentiating between different models. As an example, we cannot visually determine from the three F scores which one or two models perform better among the 5 trained machine learning models, the only thing we can say is that these 5 models are better than random guess. This is also true for Fowlkes Mallows index and Jaccard index. On the contrary, for both MCC and AUC, it is clear to observe that random forest and Gradient Boosted Machine perform better than the other three models. This is also true for balanced accuracy, bookmaker informedness and Cohen's Kappa. The general pattern is that metrics that are having a close to monotonic relationship with prevalence do a bad job distinguishing among different models while metrics that are more prevalence-independent do a better job.

We also use statistical tests to detect whether the variance in model evaluation between metrics that are more influenced by prevalence (the three F scores, Jaccard Index, TNR, TPR, and Fowlkes Mallows index) and those that are less are statistically significant. F-test, Bartlett test and Levene test results \footnote{we test the difference in variance for the interaction between metric type and model in Bartlett and Levene tests.} all show highly statistically significant p-values. That is the difference in variance of model evaluation with respect to changing prevalence is statistically significant between these two types of metrics.

\subsection{Simulation Results: Ranking of Different Models}\label{sec4.3}

\begin{table}[!htbp]
\centering
\caption{Variance of Rankings of Models by Different Metrics} 
\label{table4}
\begin{tabular}{l|r|r|r|r|r|r}
\hline
Metric & GBM & GLM & KNN & LDA & Random\_Forest & randomguess\\
\hline
True\_positives & 1.343 & 3.389 & 1.776 & 3.730 & 0.980 & 3.667\\
\hline
False\_negatives & 1.343 & 3.389 & 1.776 & 3.730 & 0.980 & 3.667\\
\hline
TPR & 1.343 & 3.389 & 1.776 & 3.730 & 0.980 & 3.667\\
\hline
F\_beta\_score\_2 & 1.321 & 3.275 & 1.913 & 3.489 & 1.196 & 3.223\\
\hline
F\_beta\_score\_0.5 & 1.404 & 2.458 & 1.780 & 2.425 & 1.649 & 2.063\\
\hline
True\_negatives & 1.360 & 2.639 & 1.242 & 3.221 & 1.169 & 2.024\\
\hline
False\_positives & 1.360 & 2.639 & 1.242 & 3.221 & 1.169 & 2.024\\
\hline
TNR & 1.360 & 2.639 & 1.242 & 3.221 & 1.169 & 2.024\\
\hline
F1\_score & 1.263 & 2.206 & 1.798 & 1.966 & 1.768 & 1.653\\
\hline
Jaccardindex & 1.263 & 2.206 & 1.798 & 1.966 & 1.768 & 1.653\\
\hline
Gmean & 1.394 & 1.207 & 1.365 & 1.103 & 1.814 & 3.188\\
\hline
Fowlkes\_Mallows & 1.354 & 1.598 & 1.502 & 1.908 & 1.287 & 0.291\\
\hline
NPV & 1.028 & 1.404 & 1.714 & 1.591 & 1.294 & 0\\
\hline
PPV & 1.183 & 1.244 & 1.402 & 1.804 & 0.745 & 0.057\\
\hline
BA & 0.755 & 0.604 & 1.310 & 0.987 & 0.944 & 0.026\\
\hline
BI & 0.755 & 0.604 & 1.310 & 0.987 & 0.944 & 0.026\\
\hline
Cohens\_kappa & 0.715 & 0.664 & 1.341 & 0.962 & 0.886 & 0.006\\
\hline
Markedness & 1.036 & 0.796 & 0.830 & 1.031 & 0.780 & 0.026\\
\hline
Diag\_odds\_ratio & 0.912 & 0.794 & 0.826 & 0.995 & 0.819 & 0.026\\
\hline
MCC & 0.588 & 0.676 & 1.251 & 0.810 & 0.714 & 0.026\\
\hline
Accuracy & 0.699 & 0.534 & 0.933 & 0.691 & 0.573 & 0\\
\hline
AUC & 0.271 & 0.419 & 0.112 & 0.309 & 0.412 & 0\\
\hline
\end{tabular}
\end{table}

\begin{figure}[!htbp]
    \caption{\textbf{Model Rankings by different metrics}}\label{fig:foobar2}
    \includegraphics[width=.49\textwidth]{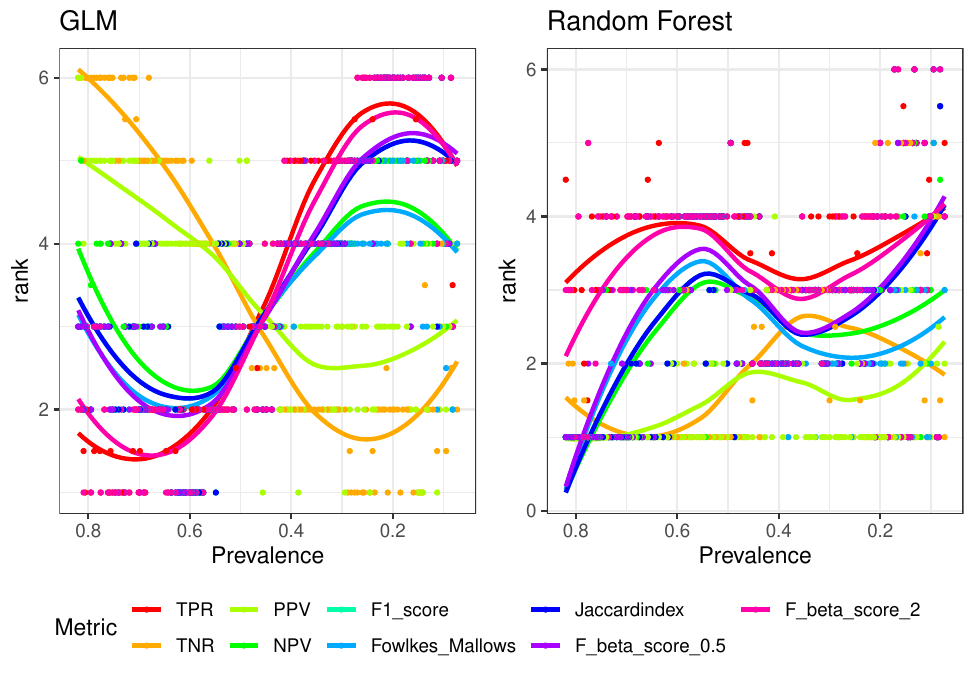}\hfill
    \includegraphics[width=.49\textwidth]{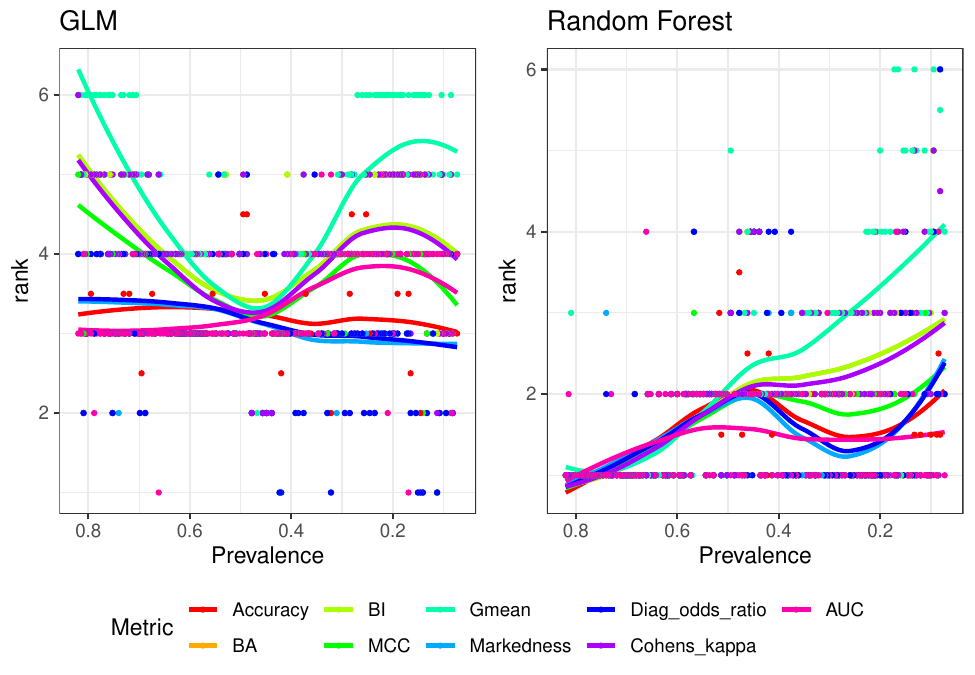}
    \includegraphics[width=.49\textwidth]{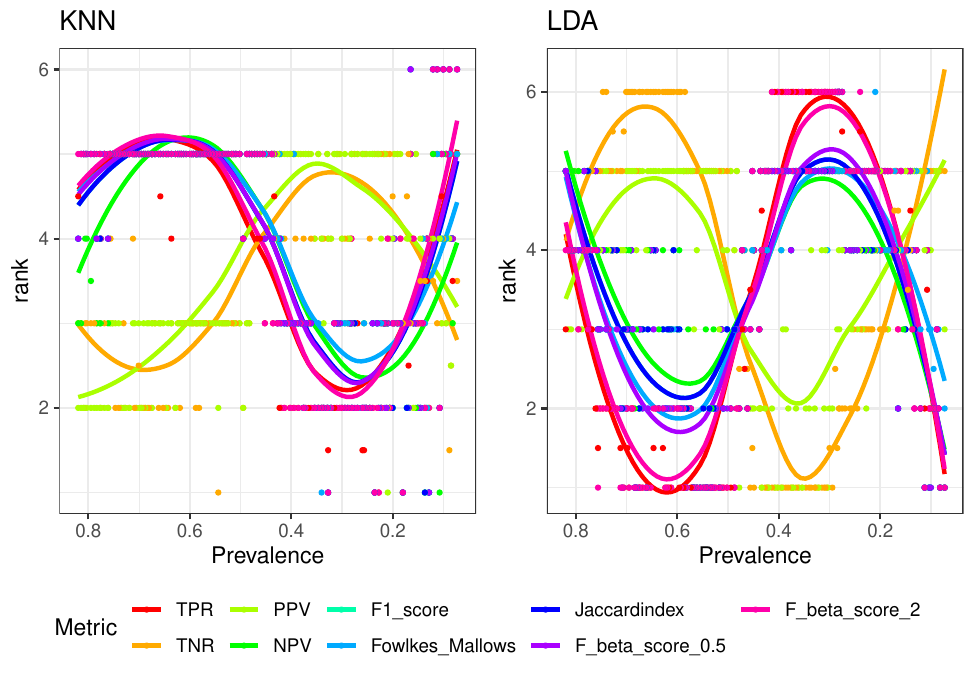}\hfill
    \includegraphics[width=.49\textwidth]{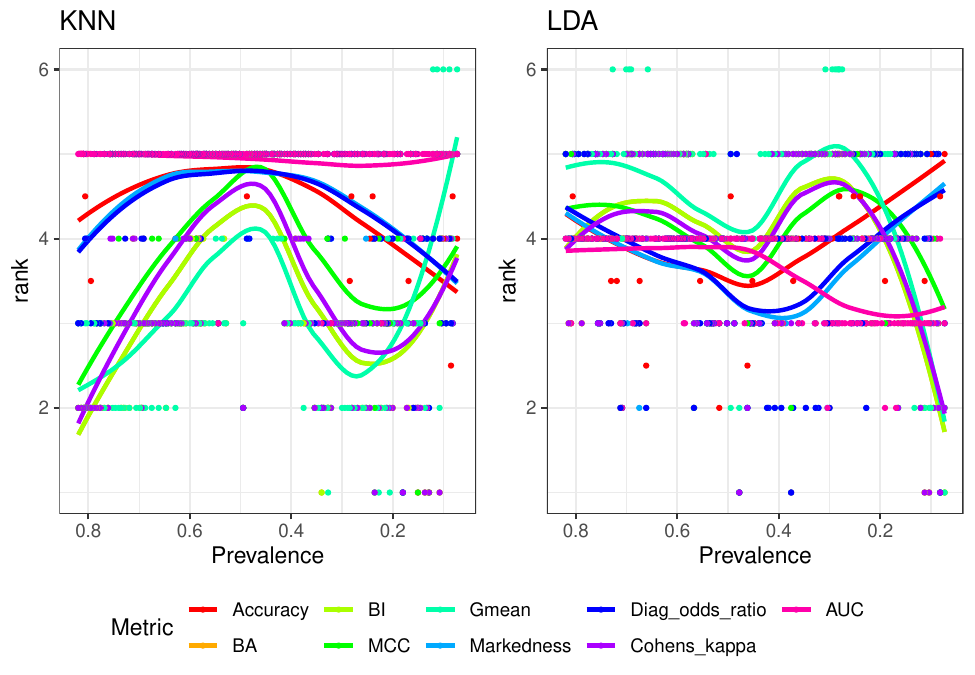}
    \includegraphics[width=.49\textwidth]{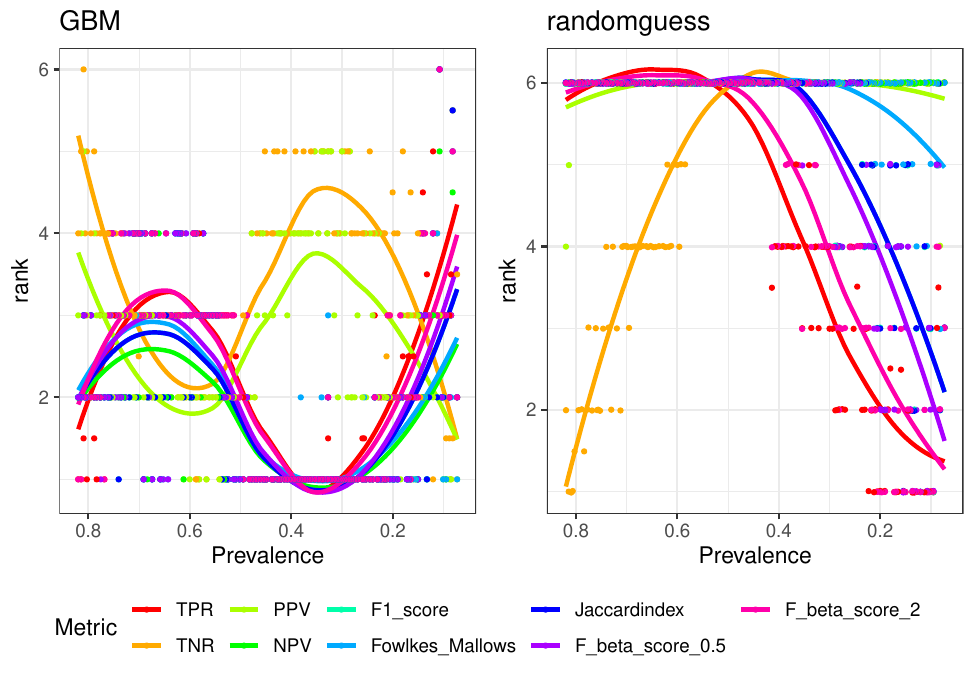}\hfill
    \includegraphics[width=.49\textwidth]{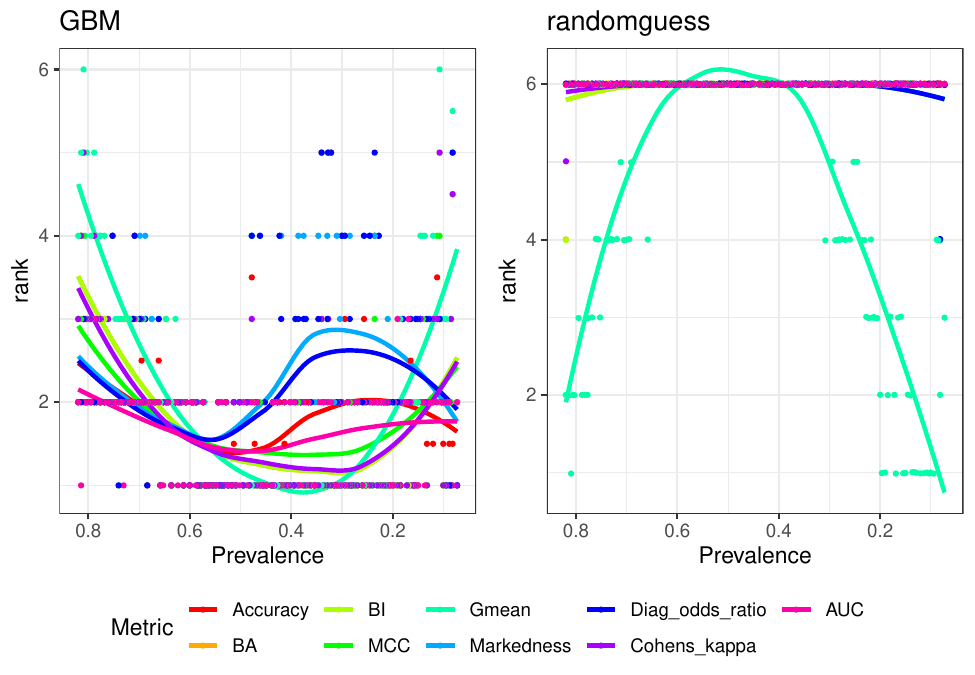}
\end{figure}

Here we provide direct results for how different metrics rank the 6 different models across data of varying prevalence, and more specifically we look at the variance of rankings for each model. Table ~\ref{table4} presents the variance values while Figure ~\ref{fig:foobar2} shows the corresponding actual rankings for each model. The variance of rankings for each metric and model combination is obtained by ranking the 6 models for each of the 156 data sets and calculating the corresponding variance, so the variance is with respect to prevalence.

Corresponding to results shown in section ~\ref{sec4.2}, it is clear to observe that metrics that are more affected by prevalence also show larger variance in ranking of different models. This type of metrics includes TPR, TNR, PPV, NPV, the three F scores, Fowlkes-Mallows index, and Jaccard index. Their variances as shown in Table ~\ref{table4} are larger compared with metrics that are less influenced by prevalence. The latter group of metrics includes BA, BI, Kappa, MCC, AUC, Markedness and DOR. Variances for the first group of metrics are almost always above 1, sometimes above 2 or 3 while variances for the second group are mostly below 1. Most notably, AUC has the smallest variance in ranking of all models as it is mostly prevalence-independent. MCC for majority of the models has the second smallest variance \footnote{The geometric mean also tries to balance prediction accuracy for both positive and negative classes yet it does have a relatively large range of values with respect to prevalence as shown in Fig ~\ref{fig:foobar1}. Correspondingly it has relatively large variance in ranking of different models.}. Figure ~\ref{fig:foobar2} shows the same patterns for actual model rankings: accuracy, balanced accuracy, bookmaker informedness, geometric mean, Kappa, MCC and AUC are shown to fluctuate less than TPR, TNR, PPV, NPV, the F-scores, Fowlkes-Mallows index, and Jaccard index.

We also perform statistical tests to detect whether the variance in model ranking between metrics that are more influenced by prevalence (the three F-scores, Jaccard Index, TNR, TPR, and Fowlkes Mallows index) and those that are less are statistically significant. F-test, Bartlett test and Levene test results \footnote{we test the difference in ranking variance for the interaction between metric type and model in Bartlett and Levene tests.} all show highly statistically significant p-values. That is the difference in variance of model rankings with respect to prevalence change is statistically significant between the two types of metrics.

\subsection{Threshold Analysis}

\begin{figure}[!htbp]
\caption{\textbf{Model Evaluation with Different thresholds at different prevalence}}\label{fig:foobar4}
\includegraphics[width=.95\textwidth]{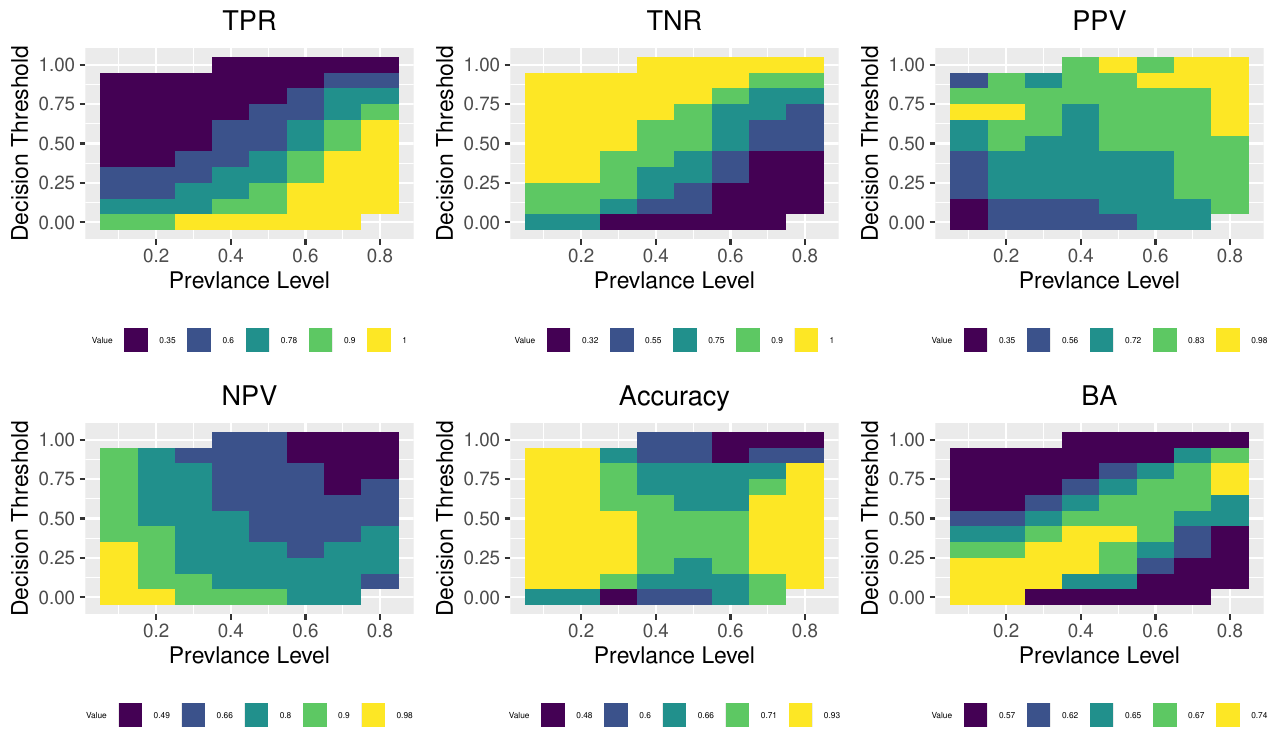}
\includegraphics[width=.95\textwidth]{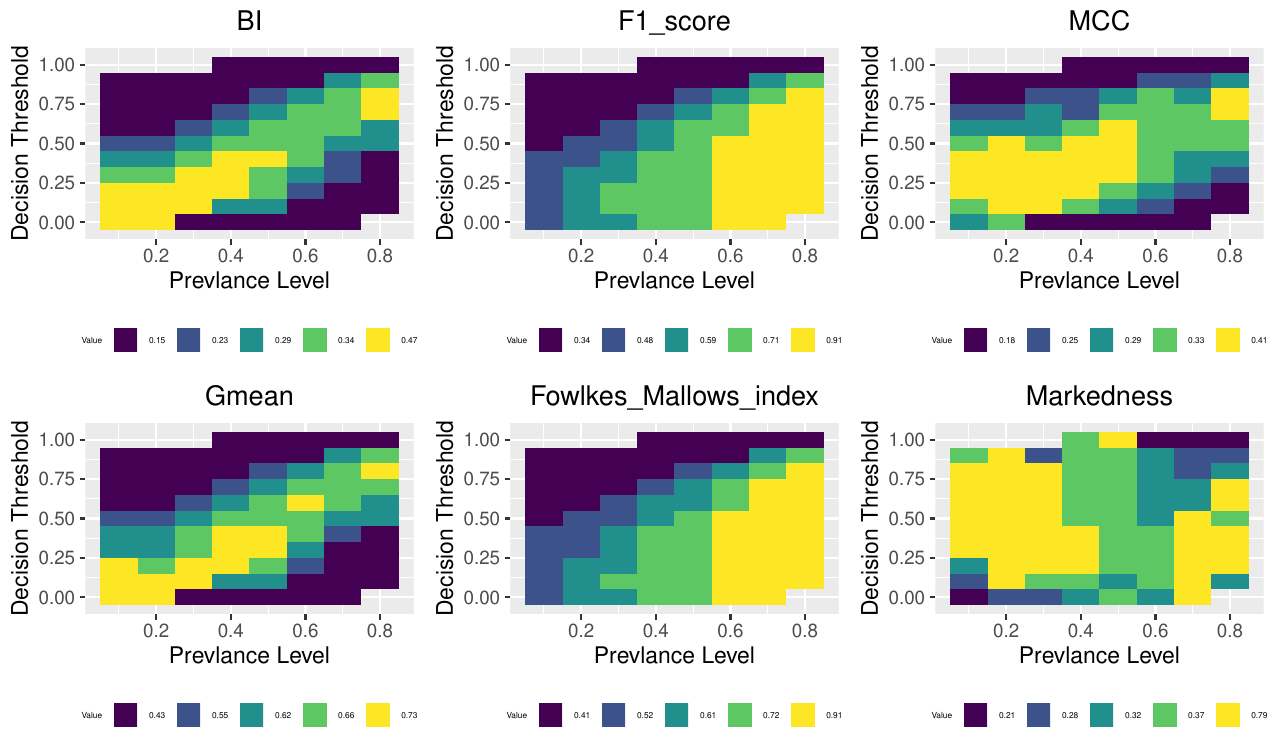}
\includegraphics[width=.95\textwidth]{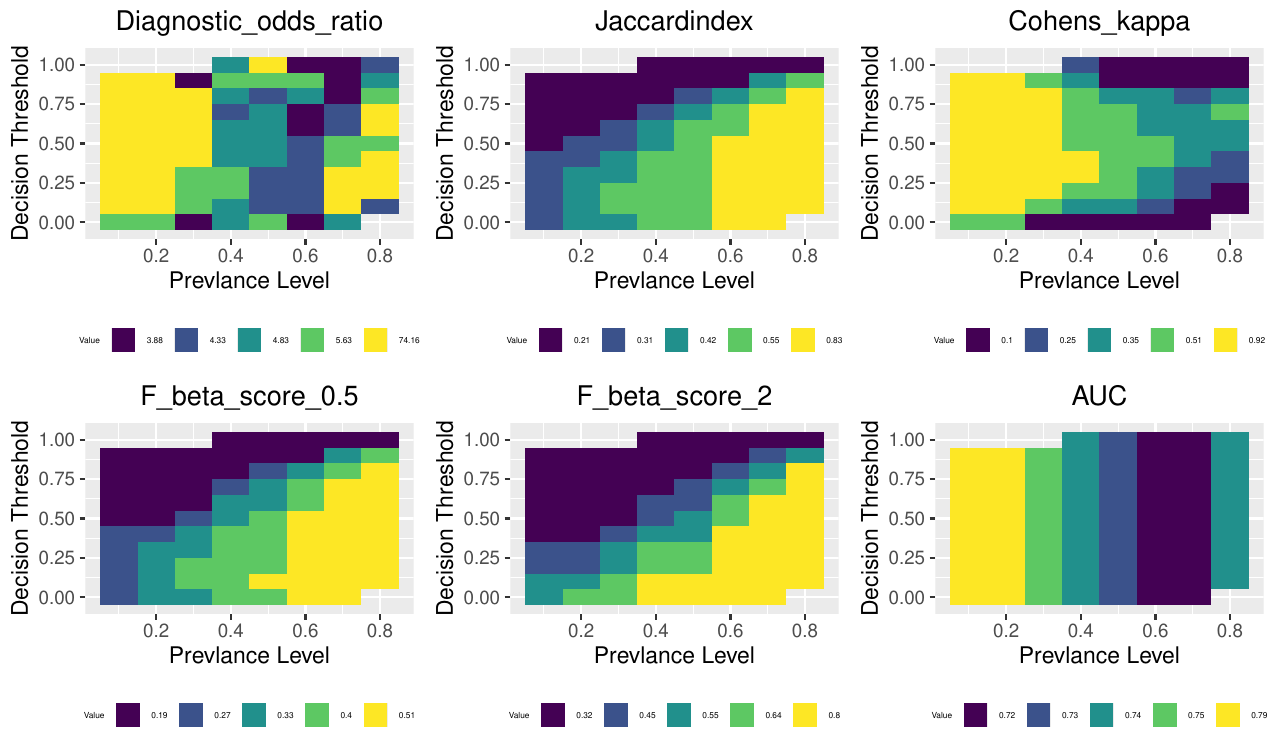}
\end{figure}

Figure ~\ref{fig:foobar4} shows the model evaluation results using different decision thresholds for data sets of different prevalence. The legend shows the upper bounds of five equally spaced intervals of the corresponding metric values with the last value being the maximum. Decision threshold refers to predicted probability for the positive class. This figure shows the results for the random forest model, results for other models show similar patterns and are included in the Supplementary Information. As can be clearly observed, metrics that are more influenced by prevalence including TPR, TNR, the three F-scores, Fowlkes Mallows Index and Jaccard Index show strong correlation with prevalence. However, this relationship is significantly moderated by the decision threshold. For example, TPR has the highest value between 0.9 and 1 when prevalence is high and threshold is low while as expected the opposite is true for TNR. The three F-scores, Fowlkes Mallows Index and Jaccard Index show similar behavior as TPR. PPV shows high values when both prevalence and decision threshold are high and low values when both are low. The opposite is true for NPV.

For metrics that try to balance accuracy for both positive and negative classes including balanced accuracy, bookmaker informedness, geometric mean, MCC, the high values reside more or less along the diagonal line of the corresponding graph while low values are away from these diagonal lines. Accuracy show high values at either high or low prevalence and low values when prevalence is close to 0.5. However, it does havve low values when the decision threshold is either close to 0 or 1 and the corresponding prevalence is low or high. As expected, AUC is invariant to decision threshold change as it always considers all possible thresholds already. And it is largely invariant to prevalence change as the range of its values is between 0.7 and 0.79, a narrower interval compared with all other metrics. Cohen's kappa has high values at low prevalence and low values at high prevalence corresponding to what is shown in Fig \ref{fig:foobar1} while there is no systematic pattern for DOR and markedness. 

These patterns correspond to what is shown in Fig \ref{fig:foobar1}, and indicates that changing from one threshold to another does have an effect on actual metric values but does not fundamentally alter the relationship between a particular metric and prevalence. In the following, we show that increasing the number of thresholds used in model evaluation does reduce metrics' variance with respect to prevalence. This provides support for the claim that the reason why AUC is able to offer more consistent model evaluation (with respect to prevalence) is because it considers all possible probability thresholds in its evaluation. 

\begin{figure}[!htbp]
    \caption{\textbf{Variance of Evaluation Metrics by Number of Thresholds.}}\label{fig:foobar5}
\includegraphics[width=.9\textwidth]{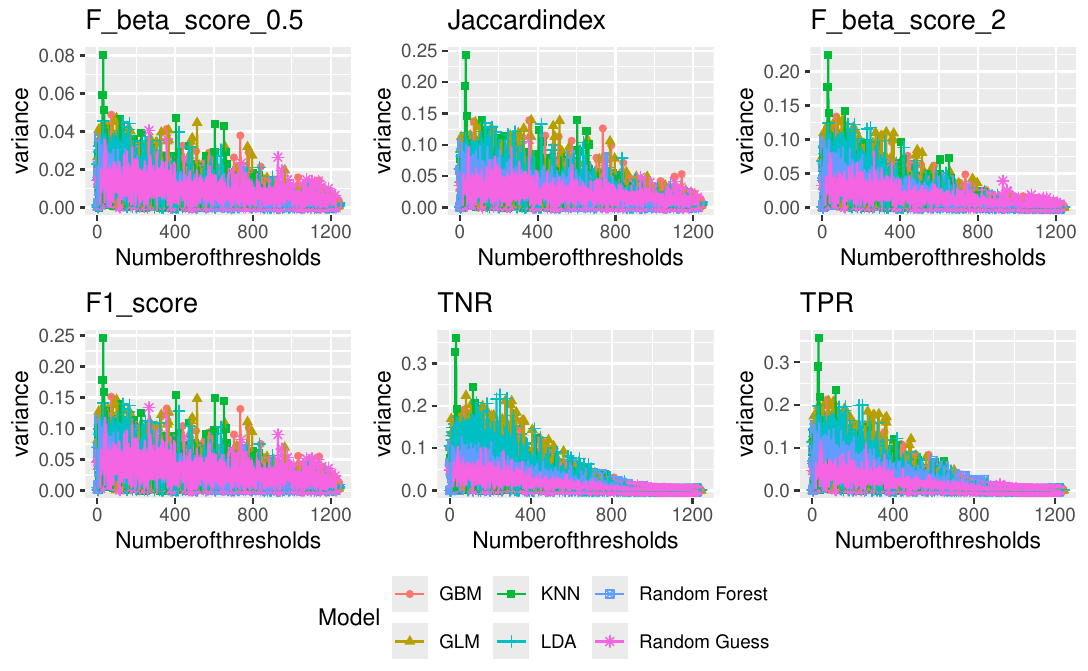}
\end{figure}

Fig \ref{fig:foobar5} shows the variance of metrics that vary most with prevalence as indicated by results shown in Fig \ref{fig:foobar1}. Here, we start with a single threshold, that is 0.5, gradually add more thresholds until all possible thresholds are added. We take the average of model evaluation metric values each time additional thresholds are added into the calculation. The thresholds are added incrementally based on their distance from 0.5. Those closest to 0.5 are added first, thus we gradually increase the span of the thresholds used. While there are fluctuations in metric values for the different models, the overall pattern is clearly that the more thresholds we consider, the smaller variance we have in regard to prevalence change.

\begin{figure}[!htbp]
     \caption{\textbf{OLS Results of Number of Thresholds on Variance of Evaluation Metrics with 95 Percent Confidence Intervals.}}\label{fig:foobar6}
\includegraphics[width=.9\textwidth]{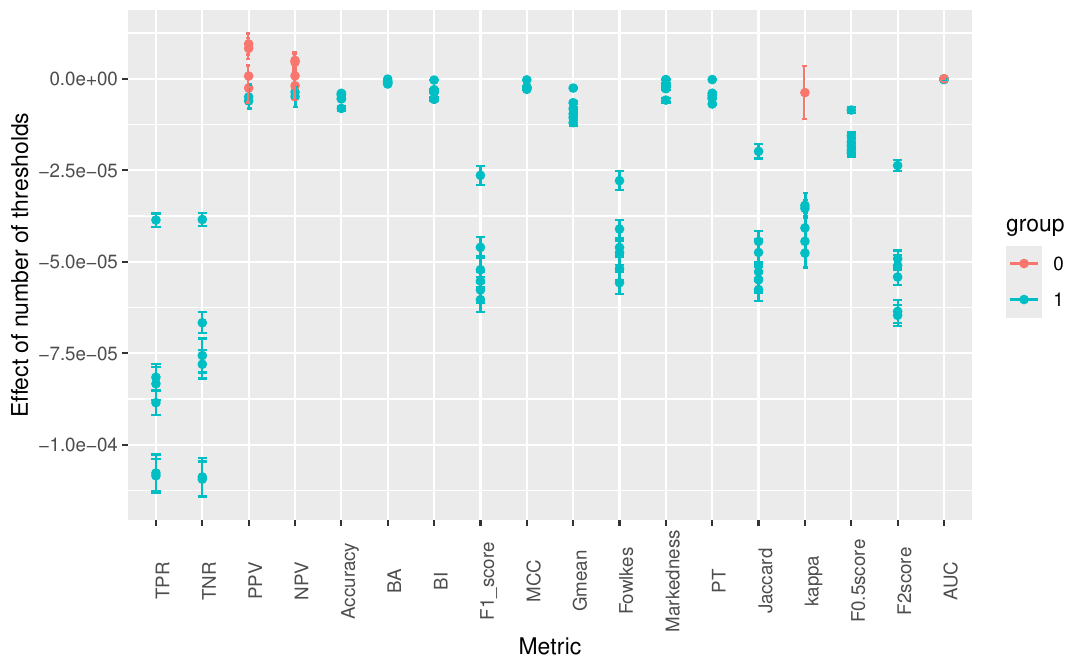}
\end{figure}

Fig \ref{fig:foobar6} shows the OLS (Ordinary Least Squares) regression results of number of thresholds used on variance of evaluation metrics. Regressing the variance of evaluation metric values on number of thresholds used in evaluation, the results show that out of 114 possible combinations of metrics and models ($19 \times 6$), 104 has a negative parameter estimate that is statistically significant (p-value less than 0.05). Compared with other metrics, DOR has a much larger negative parameter estimate that is always highly statistically significant, and is not shown here for better visualization of the other metrics. The 10 cases (in red) that either have very small non-negative parameter estimate or p-value larger than 0.05 include 8 for PPV and NPV (likely because these two metrics fluctuate so much at extreme prevalence), 1 for Cohen's Kappa (only mildly influenced by prevalence), and 1 for AUC (least influenced by prevalence and already considering all possible thresholds) \footnote{As expected the parameter estimates for AUC is always very close to zero and is not exactly zero only because the regression includes data for all models.}. The size of the parameter estimates is small because the maximum number of thresholds considered is above 1200 as shown in Fig \ref{fig:foobar5}, thus effect size of adding just one more threshold is small yet it is highly statistically significant.

\section{Discussions and Conclusion}

\subsection{Discussions}

While model evaluation for binary classification is a longstanding question with a large literature, existing research suffers from at least three shortcomings. Firstly, there is no clear attempt to establish a clean relationship between metric and prevalence level while separating out the influence of model and sample size. This paper does what is missing in the literature by simultaneously down-sampling and up-sampling different classes of data to keep both relationships among all relevant variables and sample size constant. Secondly, existing research mostly focuses on evaluating model performance on a single data set and thus ignores the question that different metrics can have systematically different behavior with respect to prevalence change. This paper addresses this question by looking at the variance of evaluation metrics across 156 data sets for 5 machine learning models as well as a naive random guess model. Lastly, existing literature still lacks a threshold analysis that can show how different metric values change as both decision threshold and prevalence change. This paper does exactly that by using all possible thresholds both individually and collectively for each metric.

The results indicate that evaluation metrics can largely be categorized into three types: monotonically influenced by prevalence such as TPR, TNR, the F-scores, Fowlkes Mallows Index and Jaccard Index; having a close to U-shaped relationship with prevalence such as balanced accuracy, bookmaker informedness, geometric mean and MCC; not having a systematic relationship with prevalence such as markedness and DOR. AUC is the only metric that is mostly prevalence-independent. Moreover, metrics that are more influenced by prevalence also have larger variance both in evaluation of individual models and in ranking of a set of models. In other words, metrics that are less influenced by prevalence offer more consistent model evaluation \footnote{The specific model being trained does have an influence, overall GBM and Random forest have smaller variance in rankings compared with GLM, KNN and LDA. As expected, random guess is most frequently being ranked as the least performing model and evaluations of almost all metrics overlap on the rank 6th line in Fig ~\ref{fig:foobar2}.}. In particular, AUC offers the most consistent model evaluation, and this is not a result by definition. The predicted probabilities for the positive and negative classes are influenced by prevalence. However, the fact that AUC evaluates model using all possible thresholds effectively offsets the influence of changing prevalence. We show evidence for this claim through a threshold analysis that includes all possible thresholds for all the data samples, models and metrics. The results show that the more decision thresholds we consider, the most consistent model evaluations we obtain across data of different prevalence. 

An implication for researchers and practitioners of model evaluation is that awareness ought to be raised for the underappreciated issue of consistency of model evaluation with respect to prevalence change. Ensuring consistency of model evaluation is important because repeated evaluation of the same model across different data samples is common. And prevalence in data collected across different time periods can change. As evaluators of model performance, we don't want metric values change just because prevalence in the data change. A much more desirable evaluation outcome is that we obtain consistent evaluation results of a model across data of different prevalence. This ensures that we are measuring the intrinsic performance of the model itself. As shown in Fig \ref{fig:foobar3}, more consistent evaluation metrics also do a better job differentiating between different models. 

As a practical guide for action, if practitioners are interested in the complete picture of model performance for a binary classification task, then they need to look at all four basic quantities (TP, TN, FN, FP). If practitioners are interested in one single metric to summarize the model performance, then the particular goal of the specific evaluation task must be taken into consideration as no single metric can summarize the four basic quantities satisfactorily in general. If the goal of evaluation is to obtain consistent rating of a single model or consistent ranking of a set of models across different data scenarios, then AUC should be the metric to be used. 

One limitation of this study is that the simulation exercises performed above focuses on one data set which has a more or less balanced outcome variable, similar analysis with other data sets with a more imbalanced outcome can be performed to further confirm the current findings and we expect similar patterns to hold. Moreover, in training the five models above, seven features are used as predictor variables. In other settings of applied machine learning, the number of predictor variables can be much larger and that can be an additional source of variation to a prediction task. Further analysis of data with a much larger set of feature variables can potentially bring additional insights to the complex interplay of data, model and model evaluation metrics.

Regarding the implications for different domains or types of data, no definite answer can be given at this point. However, we expect similar results to hold in settings with a much large sample size and a lot more feature variables. The number of feature variables do not have an effect on evaluation metric values as long as the relationships between the feature variables and the outcome variable are kept constant and the same model is used. Likewise, sample size alone should not have an effect on evaluation metrics as sample size is not in the formulation of almost all metrics as is shown in section \ref{sec2.1}. When the number of variables is higher than sample size, we are in a high-dimensional setting and things can become different as a different modeling strategy is needed. 

Another limitation of this study is that we only consider binary classification tasks, similar analysis for data sets with multi-class outcome can be conducted. For a multi-class classification question, often we need to binarize the outcome in certain ways: either adopting a one-versus-rest approach where one class is compared against all the other classes or taking a one-versus-one approach where comparison is made for every pairwise class combination. And the formulation for most metrics can be easily extended to a multi-class scenario. This will enable generalizing the current study to the multi-class setting. For a one-versus-rest comparison, we expect similar results to hold, while for one-versus-one pairwise comparison, we expect similar results to hold conditional on the relative size of other classes not in the comparison.

\subsection{Conclusion}

In conclusion, this study provides new analytical results on machine learning model evaluation for binary classification and more specifically for the consistency of model evaluation metrics with respect to prevalence change while holding the influence of model and data constant. The analysis performed covers a distribution of data scenarios, from an overwhelming positive sample to an overwhelming negative sample and everything in between. Across this full span of data scenarios and five commonly used machine learning models as well as naive random guess, clear patterns are obtained for 18 evaluation metrics. 

It is hoped that the results presented above can provide a different angle on thinking about the complex relationships between data, model and evaluation metrics and can spur further research into developing best practices for machine learning model evaluation in various scientific domains. As has been shown above, statistical simulation can be an effective tool for scientific discovery even in an age of big data. Only analyzing observed empirical data is not enough for full-scale scientific advances. Theory-driven simulation analysis can provide new findings that are not available otherwise.  

\section*{Supporting information}

Included are model evaluation heat maps using different decision thresholds and correlation heat maps that showcase the bivariate correlations among the predictors and the outcome for the original data sample as well as 79 simulated data samples (every other ones selected from all the data samples).

\section*{Appendices}
\appendix

\section{Prediction Results for Crime Recidivism}\label{appendix1}

\begin{table}[!htbp]
\centering
\caption{Crime Recidivism Model Evaluation Metric Values} 
\label{table3}
\begin{tabular}{l|r|r|r|r|r|r}
\hline
Metric & GBM & GLM & KNN & LDA & Random Forest & randomguess\\
\hline
True\_positives & 339 & 292 & 296 & 286 & 311 & 267\\
\hline
False\_negatives & 223 & 270 & 266 & 276 & 251 & 295\\
\hline
True\_negatives & 517 & 537 & 521 & 543 & 543 & 386\\
\hline
False\_positives & 164 & 144 & 160 & 138 & 138 & 295\\
\hline
TPR & 0.603 & 0.520 & 0.527 & 0.509 & 0.553 & 0.475\\
\hline
TNR & 0.759 & 0.789 & 0.765 & 0.797 & 0.797 & 0.567\\
\hline
PPV & 0.674 & 0.670 & 0.649 & 0.675 & 0.693 & 0.475\\
\hline
NPV & 0.699 & 0.665 & 0.662 & 0.663 & 0.684 & 0.567\\
\hline
Accuracy & 0.689 & 0.667 & 0.657 & 0.667 & 0.687 & 0.525\\
\hline
BA & 0.681 & 0.654 & 0.646 & 0.653 & 0.675 & 0.521\\
\hline
BI & 0.362 & 0.308 & 0.292 & 0.306 & 0.351 & 0.042\\
\hline
F1\_score & 0.637 & 0.585 & 0.582 & 0.580 & 0.615 & 0.475\\
\hline
MCC & 0.367 & 0.321 & 0.301 & 0.322 & 0.363 & 0.042\\
\hline
Gmean & 0.677 & 0.640 & 0.635 & 0.637 & 0.664 & 0.519\\
\hline
Fowlkes\_Mallows & 0.638 & 0.590 & 0.585 & 0.586 & 0.619 & 0.475\\
\hline
Markedness & 0.373 & 0.335 & 0.311 & 0.338 & 0.377 & 0.042\\
\hline
Diag\_odds\_ratio & 4.792 & 4.033 & 3.623 & 4.077 & 4.875 & 1.184\\
\hline
Jaccardindex & 0.467 & 0.414 & 0.410 & 0.409 & 0.444 & 0.312\\
\hline
Cohens\_kappa & 0.366 & 0.314 & 0.297 & 0.313 & 0.357 & 0.042\\
\hline
F\_beta\_score\_0.5 & 0.352 & 0.321 & 0.319 & 0.317 & 0.338 & 0.264\\
\hline
F\_beta\_score\_2 & 0.512 & 0.450 & 0.453 & 0.442 & 0.477 & 0.396\\
\hline
AUC & 0.734 & 0.718 & 0.694 & 0.717 & 0.727 & 0.524\\
\hline
\end{tabular}
\end{table}

Table \ref{table3} shows the crime recidivism prediction results for 5 machine learning models (logistic regression (GLM), random forest (RF), k-nearest neighbors (KNN), linear discriminant analysis (LDA), gradient boosting machine (GBM)) as well as naive random guess evaluated with 18 metrics. The data set \citep{bansak_can_2019} consists of individuals arrested in Broward County, Florida between 2013 and 2014. The sample size is 6214, with 2775 positive cases, individuals who reoffended and 3439 negatives, individuals who did not reoffend. The predictors for the classifiers include seven features of the defendants: gender, age, number of juvenile misdemeanors, number of juvenile felonies, number of prior (nonjuvenile) crimes, crime degree, and crime charge. 

\section{F1 score as a result of prevalence change when only sample size n is constant}\label{appendix3}

From the alternative definition of $\text{F1 score}$ in equation \ref{eq:2}, we know that $\text{F1 score} = \frac{2} {2  + \frac{1}{\text{TPR}}((1 - \text{TNR}) \frac {1 - \phi}{\phi}  + 1 - \text{TPR})}$, in addition, $\phi ~ \text{decreases}, \text{TPR} ~ \text{decreases}, \text{TNR} ~ \text{increases} \quad  \Rightarrow \quad \frac{1}{\text{TPR}} ~ \text{increases} , (1 - \text{TNR}) ~ \text{decreases}$, $\frac {1 - \phi}{\phi} ~ \text{increases}$, and $(1 - \text{TPR}) ~ \text{increases}$. And across the whole spectrum of prevalence, the rate of change in $\text{TPR}$ is similar to the rate of change in both $\text{TNR}$ and prevalence itself, and as only $(1 - \text{TNR})$ goes down while all other terms in the denominator of F1 score goes up, we have that $((1 - \text{TNR}) \frac {1 - \phi}{\phi}  + 1 - \text{TPR} ) ~ \text{increases}$ and $\text{F1 score} ~ \text{decreases}$ as prevalence level decreases. 

\bibliography{refs}

\newpage
\section{Supplementary Information}

For all models, in certain small regions of the following heat maps, there is change in value for AUC when evaluating using different thresholds, this is just an artifact of plotting. The reason is because there are more than one prevalence level at that point of the x-axis, and they happen to have different AUC. AUC as expected should be completely independent of the specific threshold chosen as it already considers all possible thresholds.

\begin{figure}[!htbp]
\caption{Model Evaluation for different prevalence and decision thresholds for GLM.}
    \includegraphics[width=.5\textwidth]{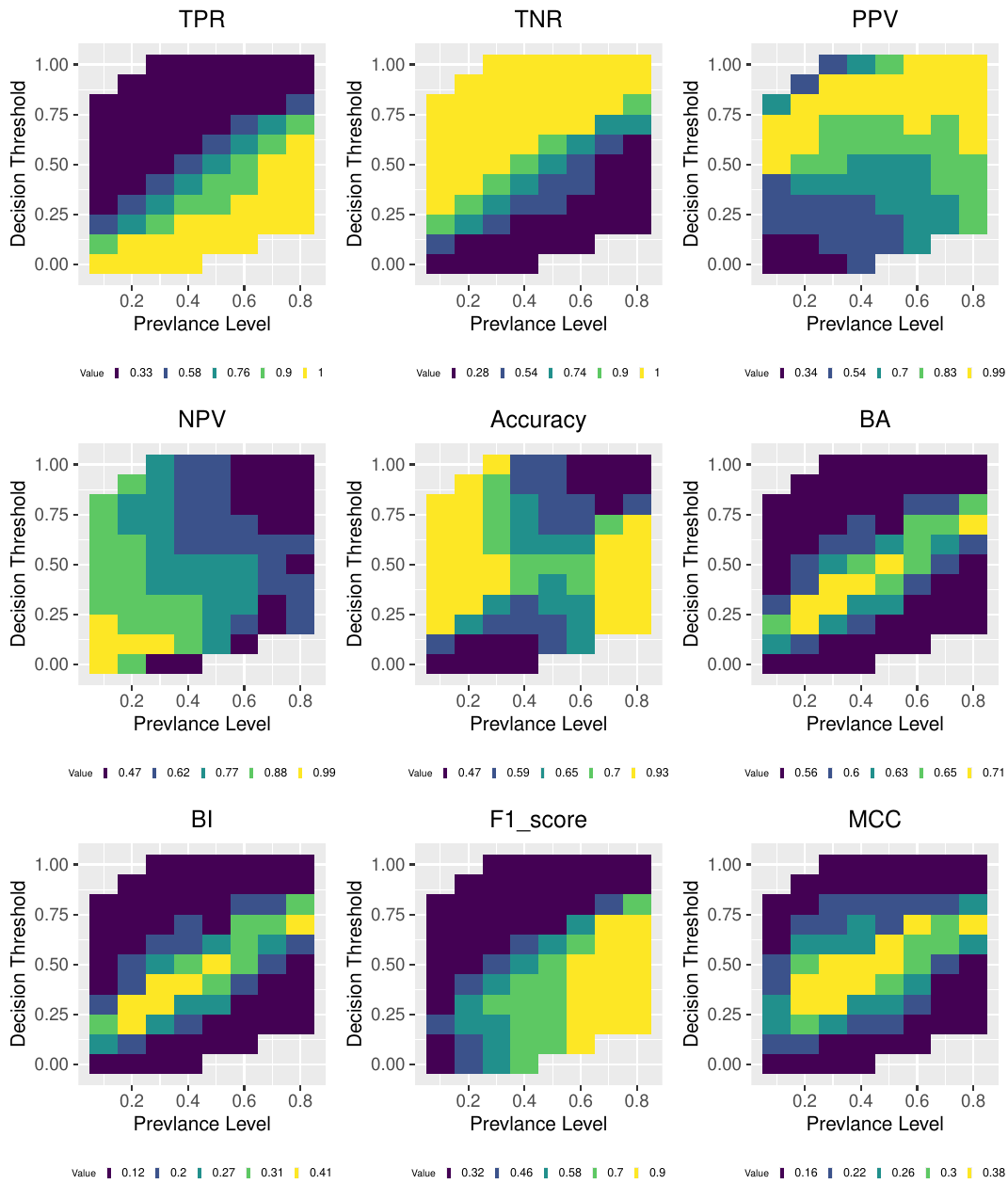} 
        \includegraphics[width=.5\textwidth]{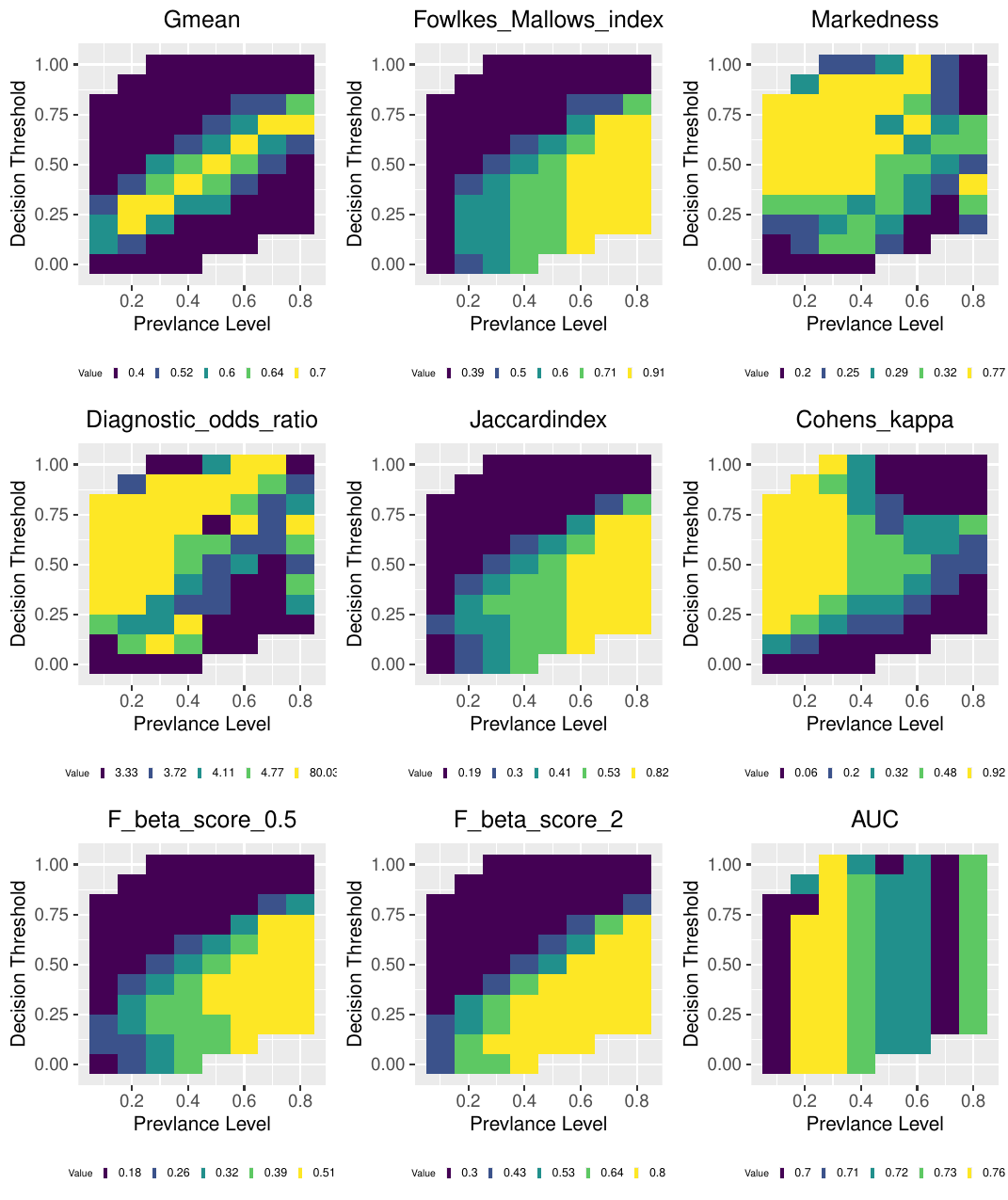} 
\end{figure}

\begin{figure}[!htbp]
\caption{Model Evaluation for different prevalence and decision thresholds for KNN}
    \includegraphics[width=.5\textwidth]{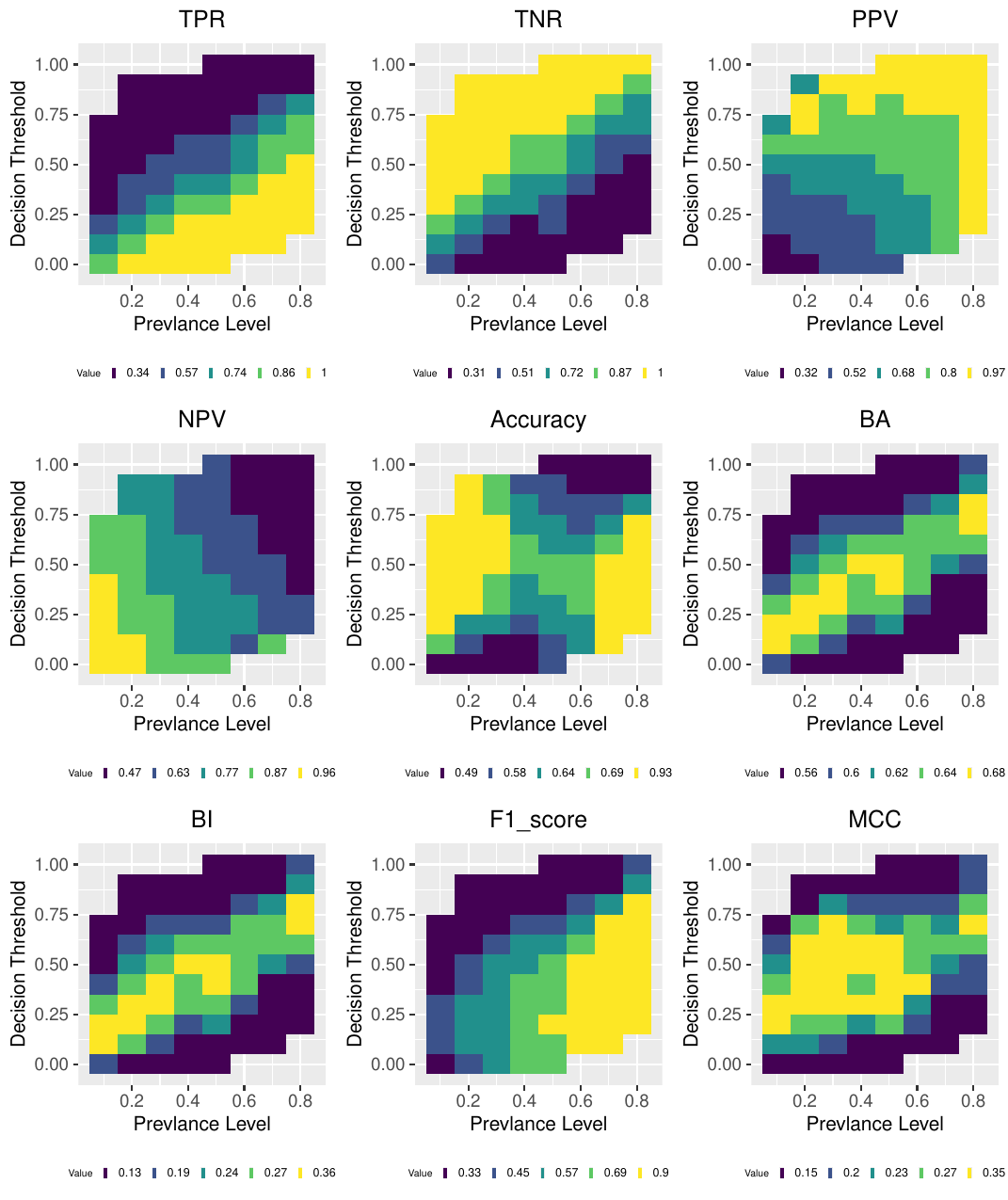} 
        \includegraphics[width=.5\textwidth]{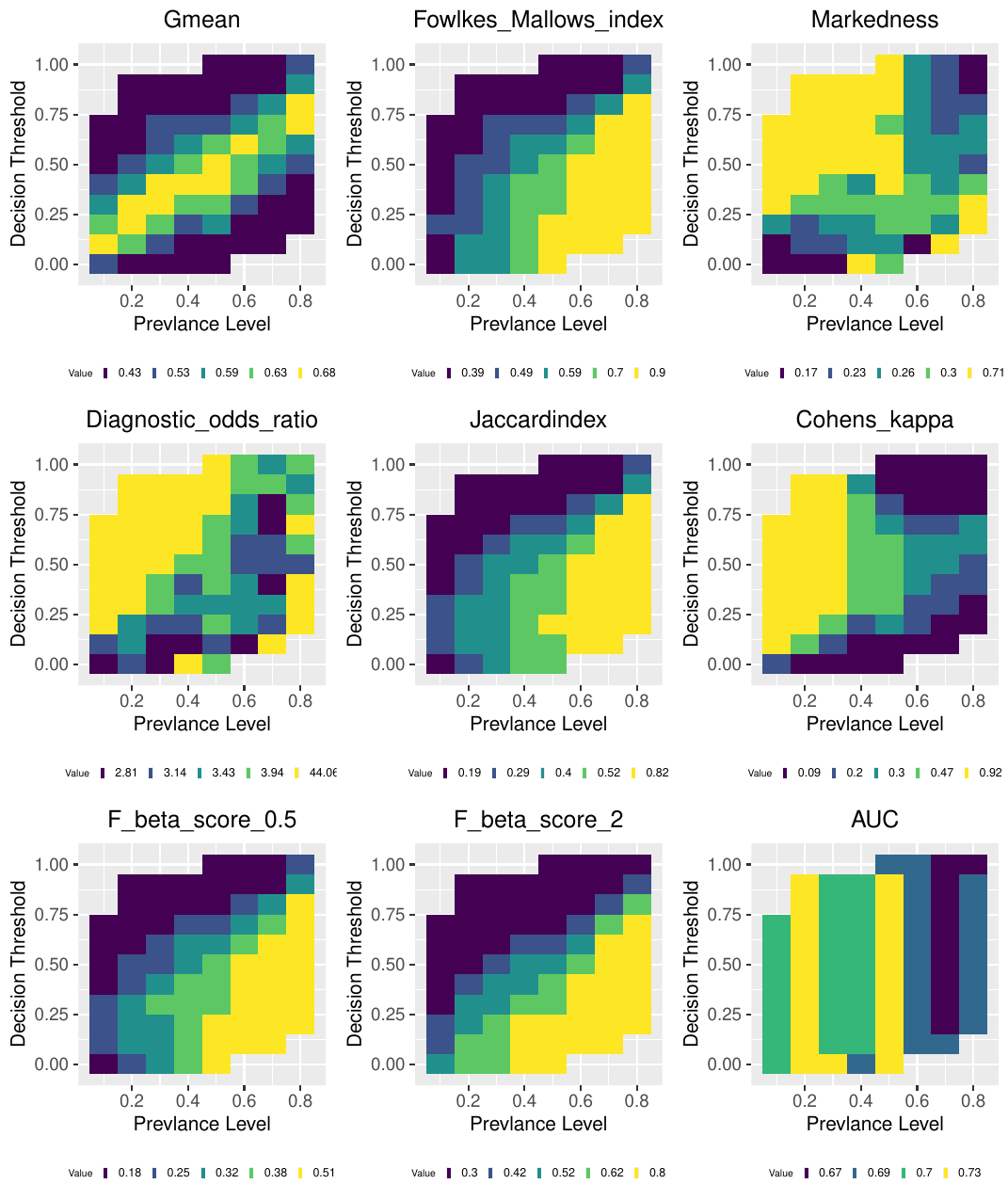} 
\end{figure}

\begin{figure}[!htbp]
\caption{Model Evaluation for different prevalence and decision thresholds for LDA}
    \includegraphics[width=.5\textwidth]{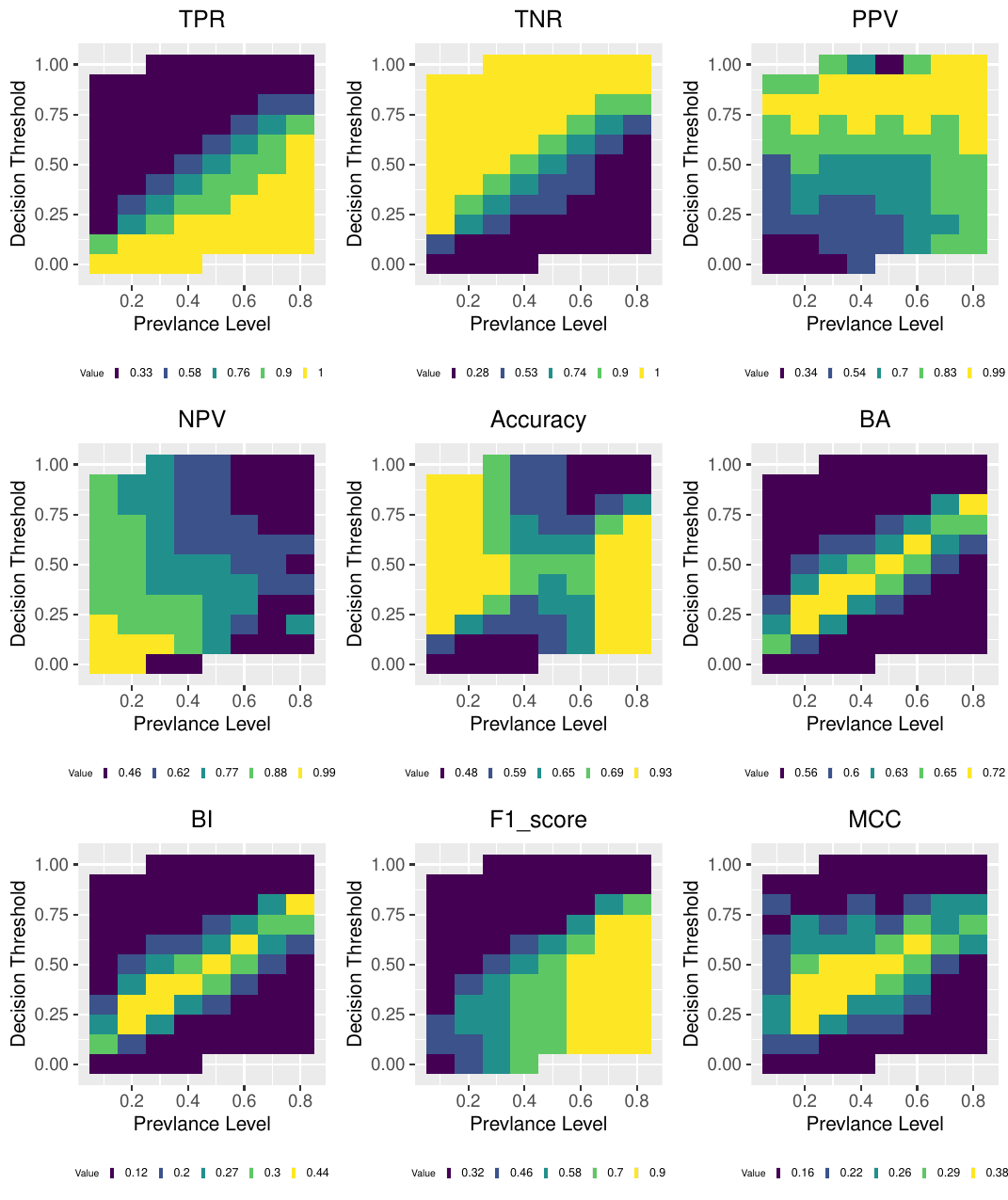} 
        \includegraphics[width=.5\textwidth]{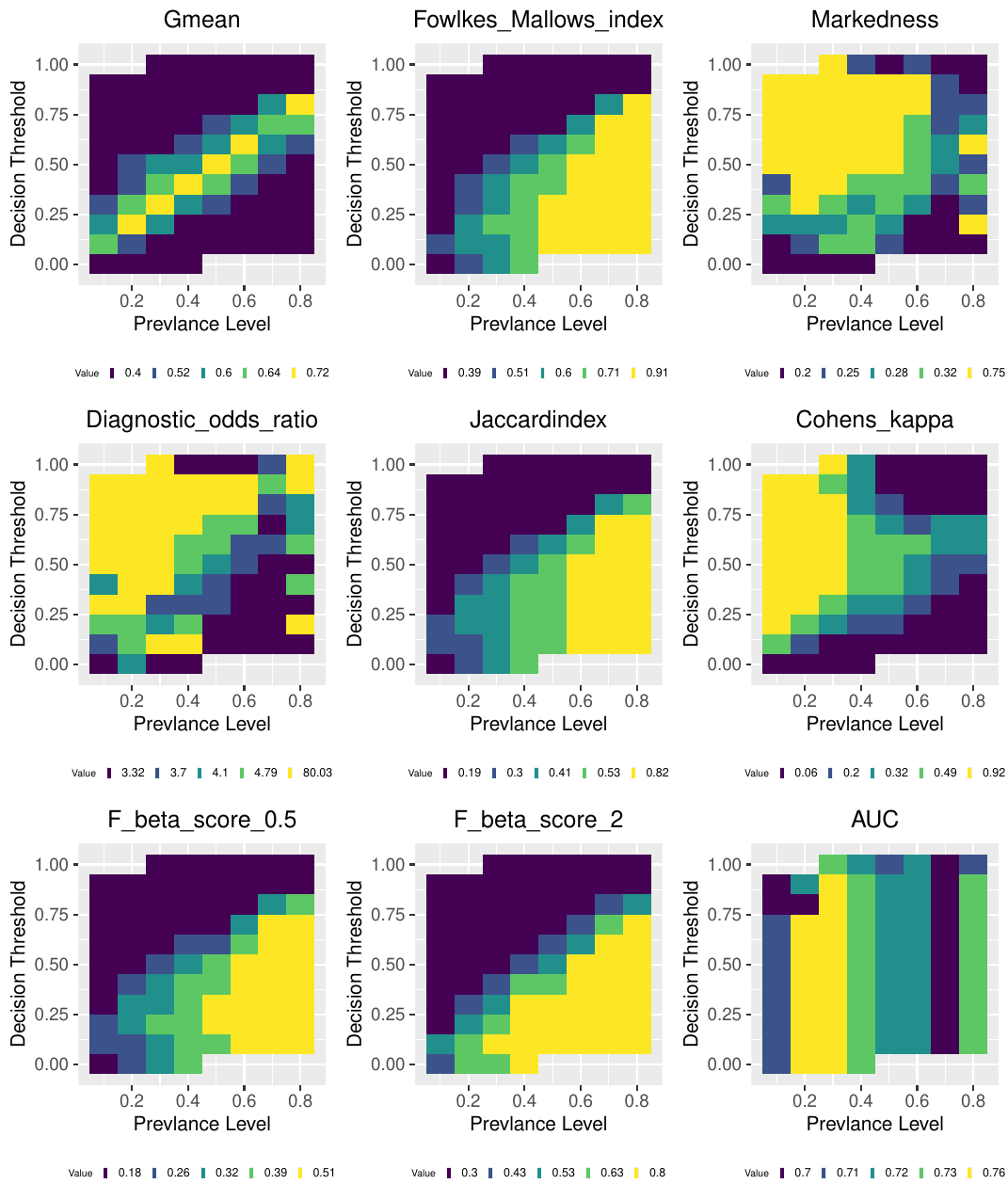} 
\end{figure}

\begin{figure}[!htbp]
\caption{Model Evaluation for different prevalence and decision thresholds for GBM}
    \includegraphics[width=.5\textwidth]{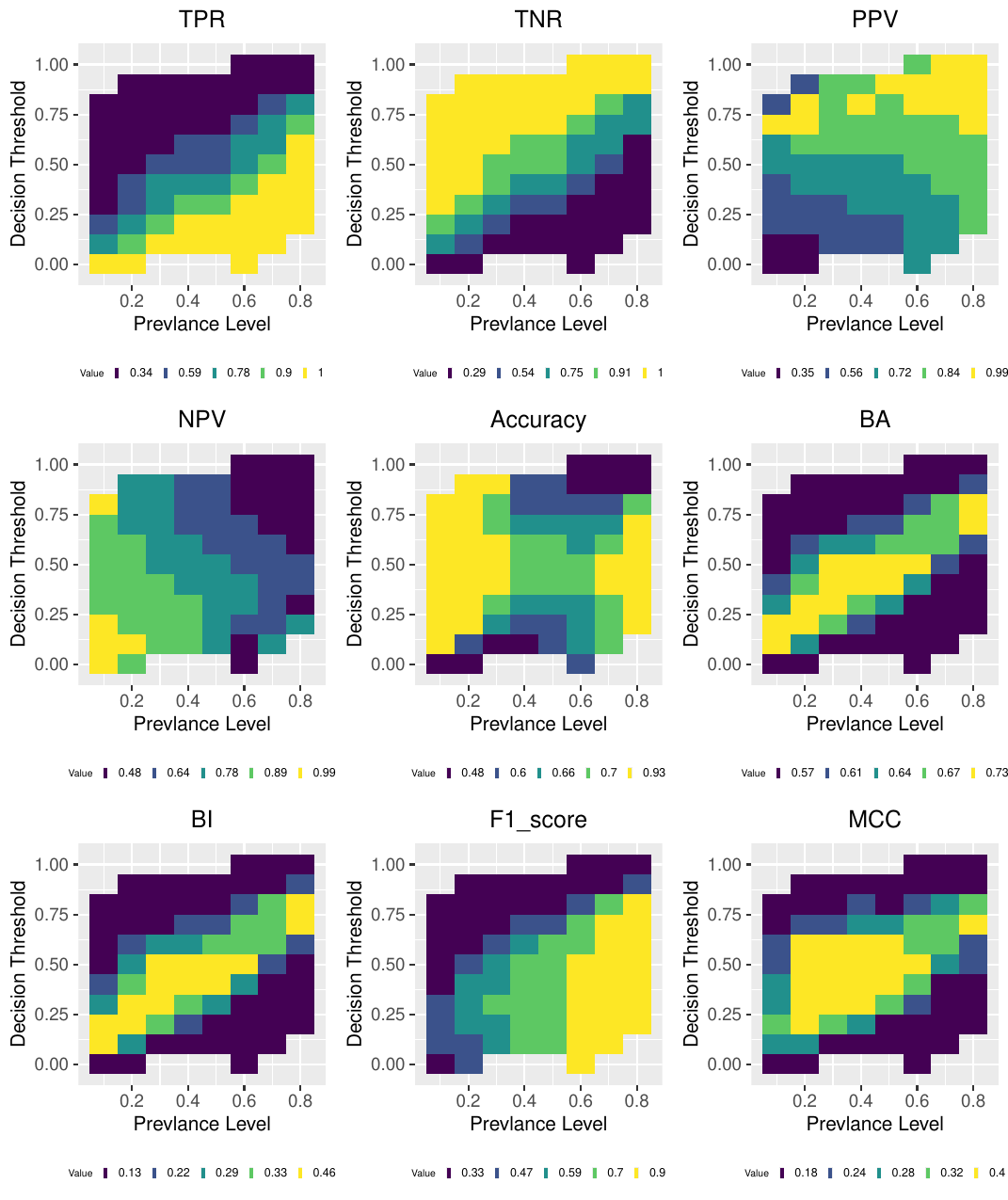} 
        \includegraphics[width=.5\textwidth]{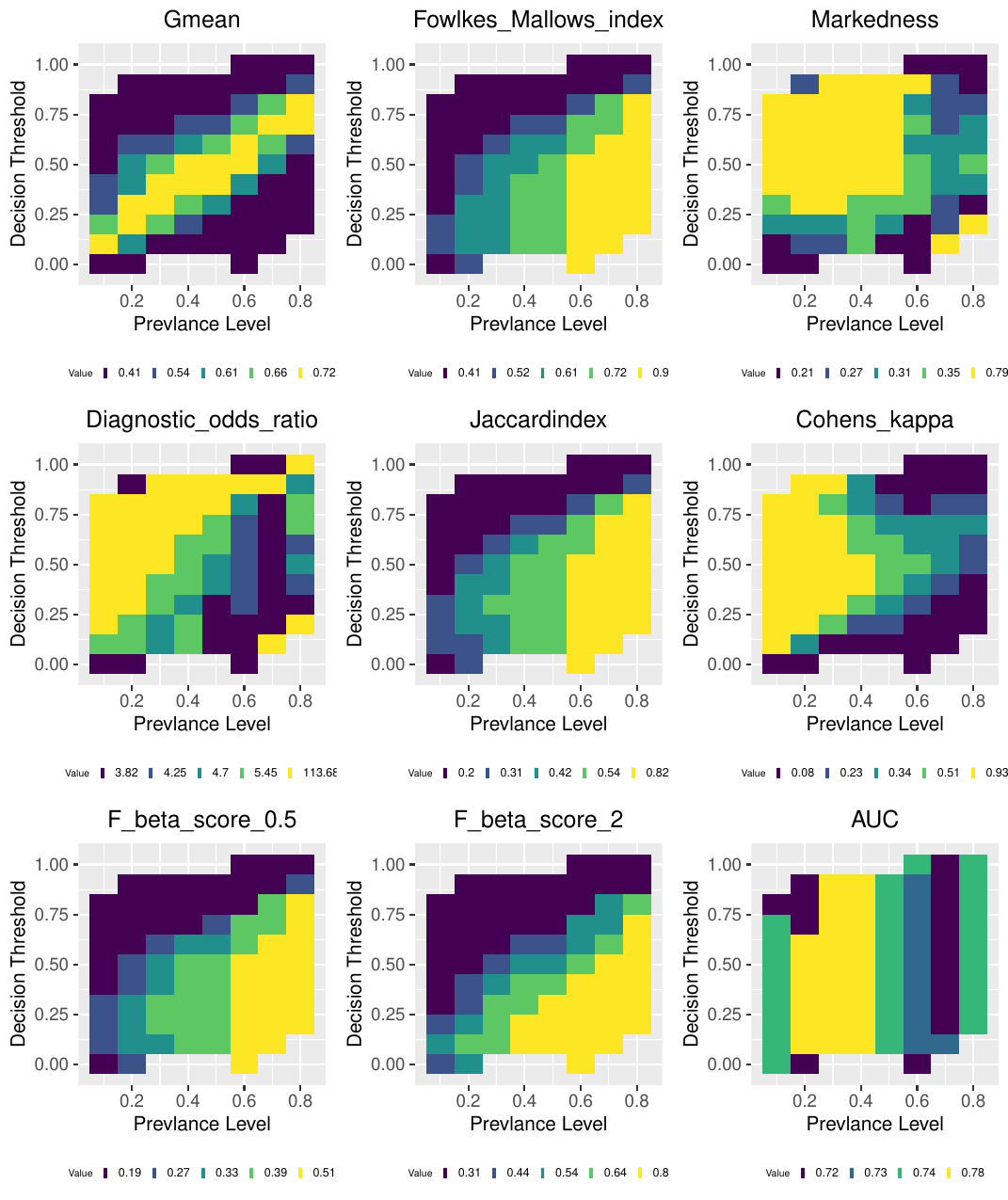} 
\end{figure}

\begin{figure}[!htbp]
\caption{Model Evaluation for different prevalence and decision thresholds for Random Guess.}
    \includegraphics[width=.5\textwidth]{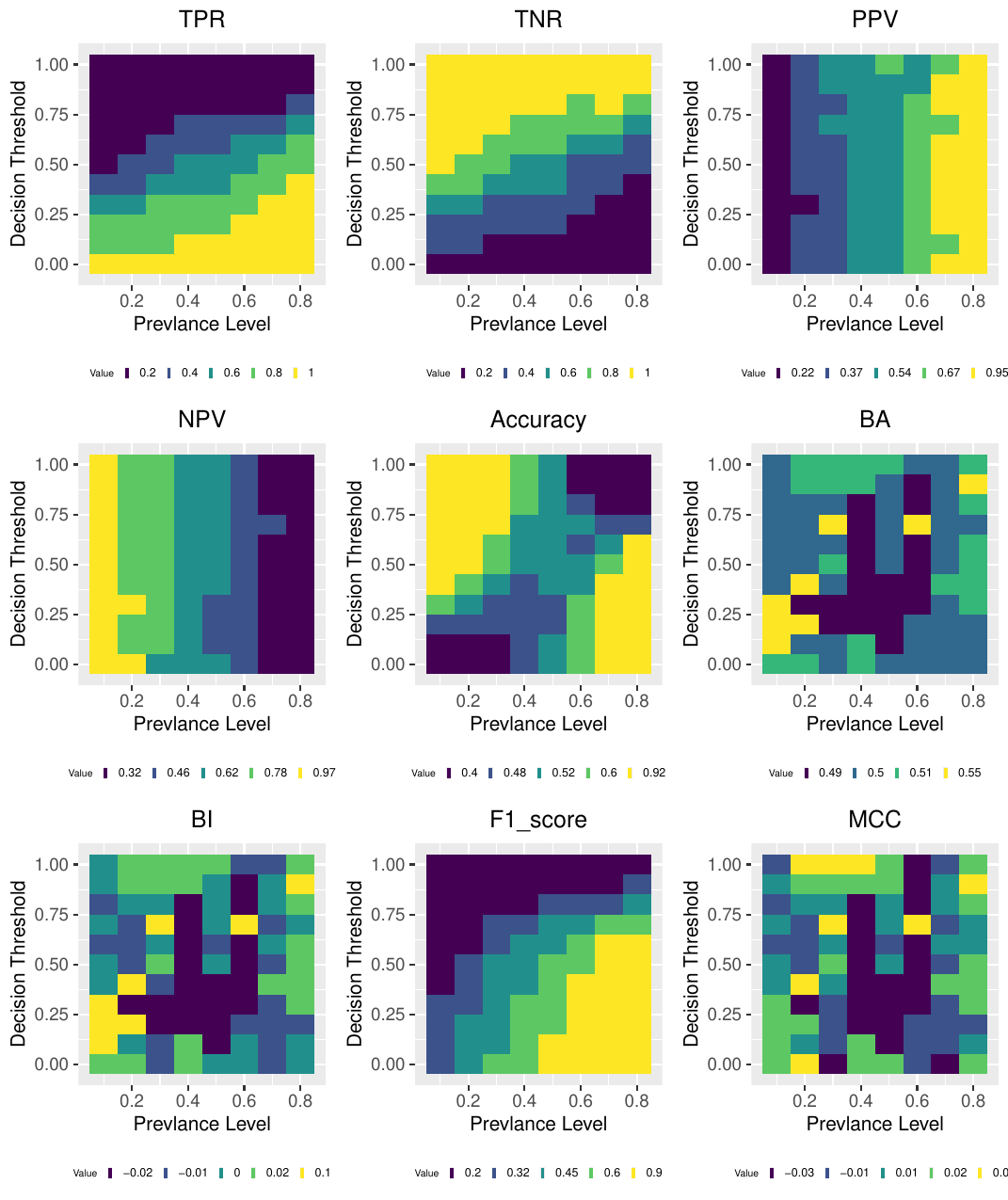} 
        \includegraphics[width=.5\textwidth]{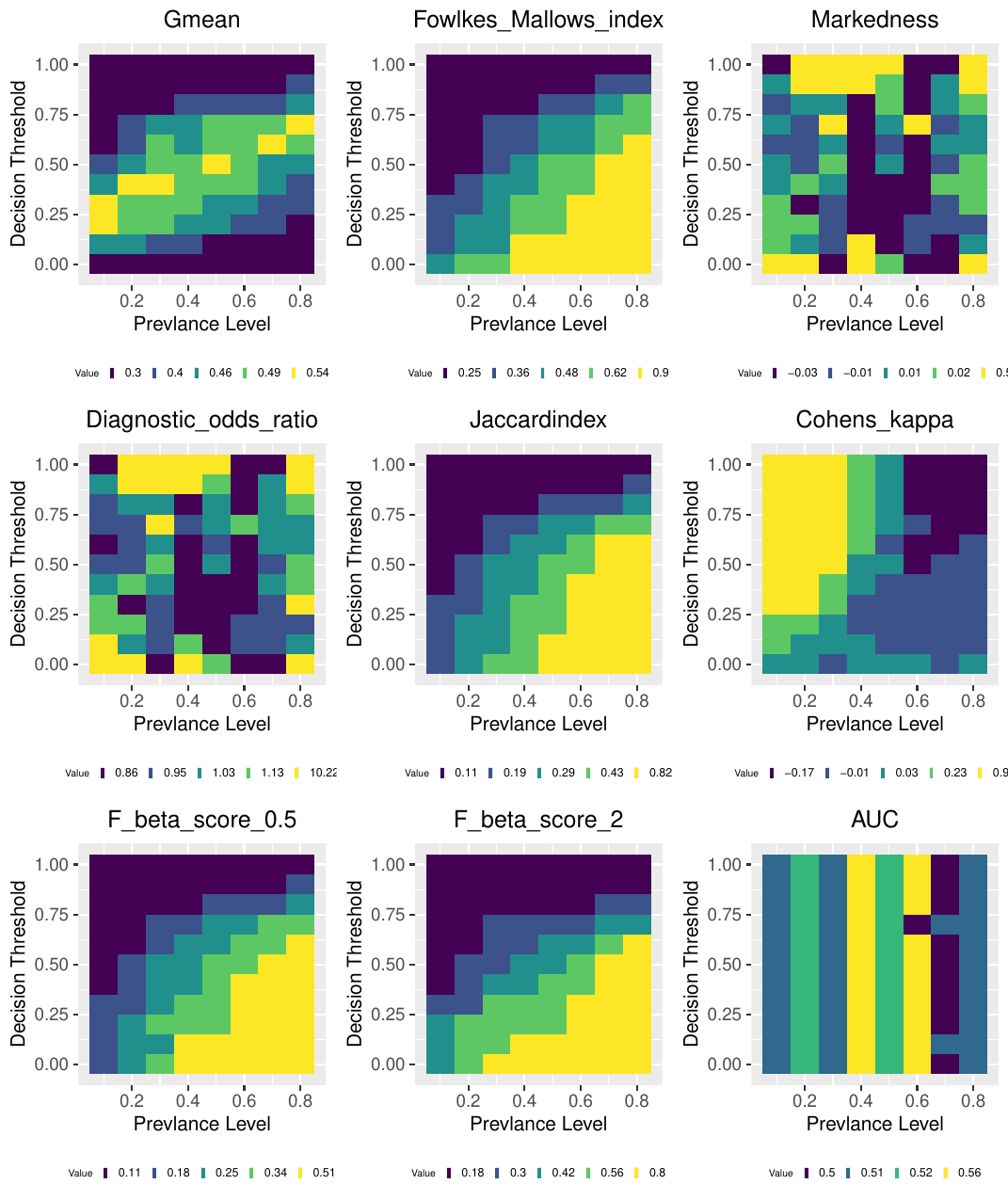} 
\end{figure}

\begin{figure}
\caption{Correlation Heat Maps for Different Data Samples. Labels for each pane indicates the random number of positive cases being added or dropped from the original data set.}
\includegraphics[width=.5\textwidth]{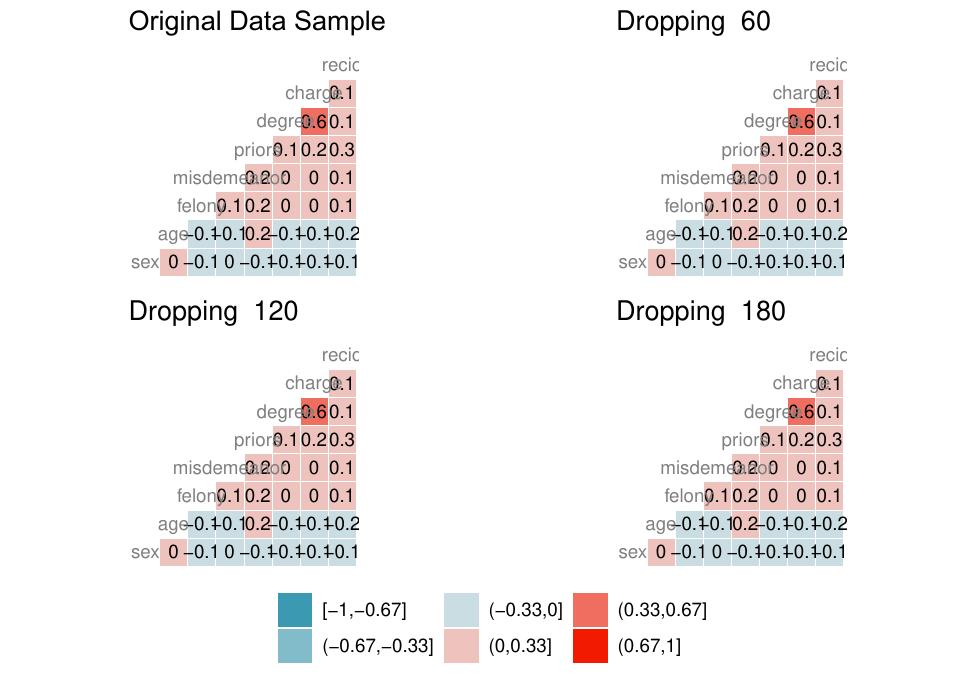}
\includegraphics[width=.5\textwidth]{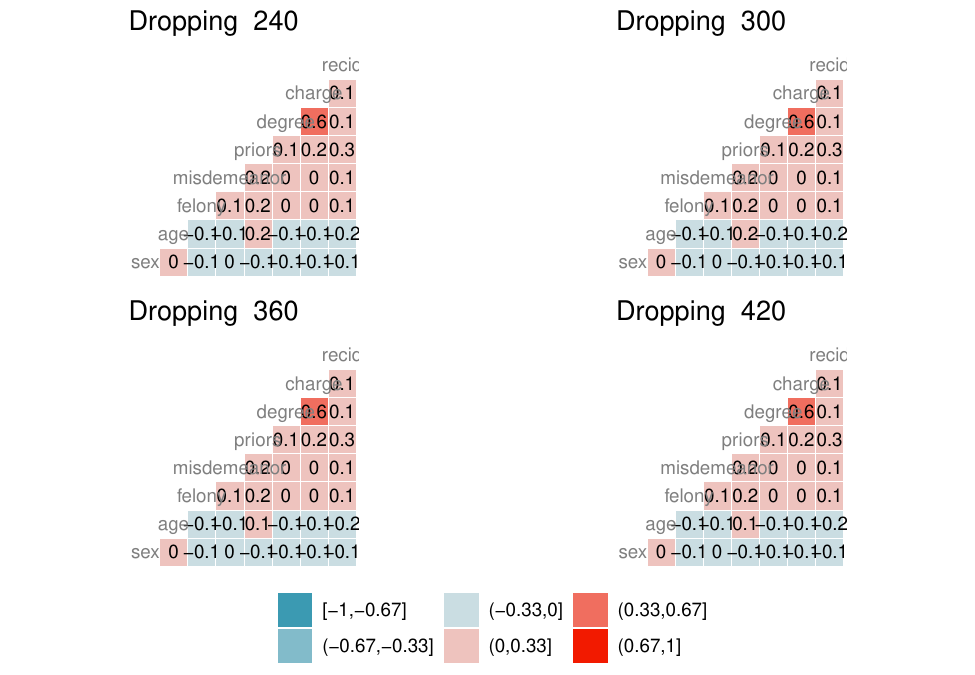}
\includegraphics[width=.5\textwidth]{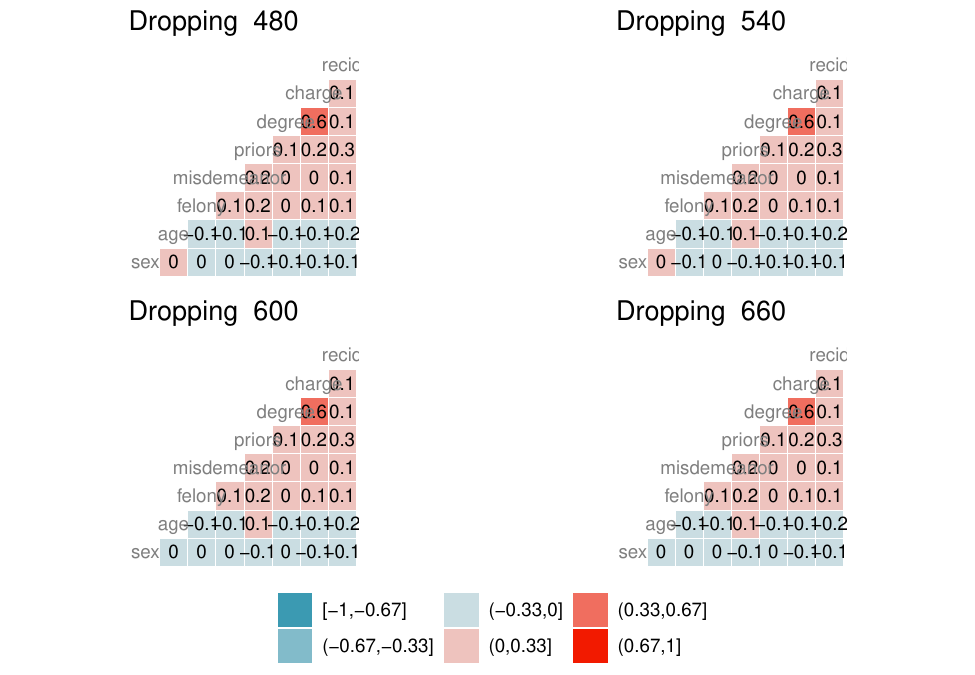}
\includegraphics[width=.5\textwidth]{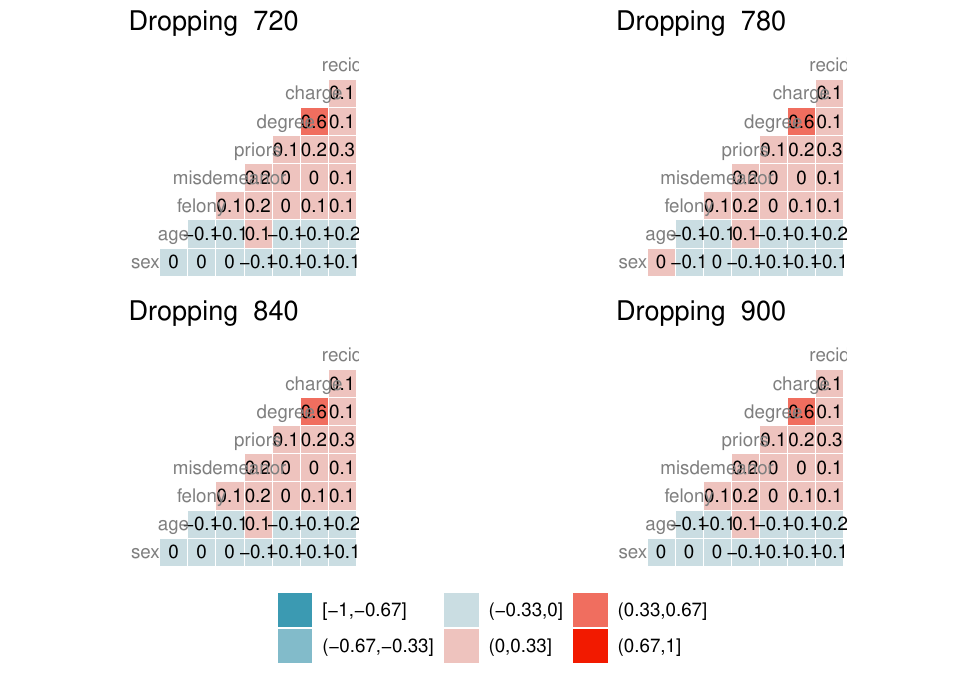}
\includegraphics[width=.5\textwidth]{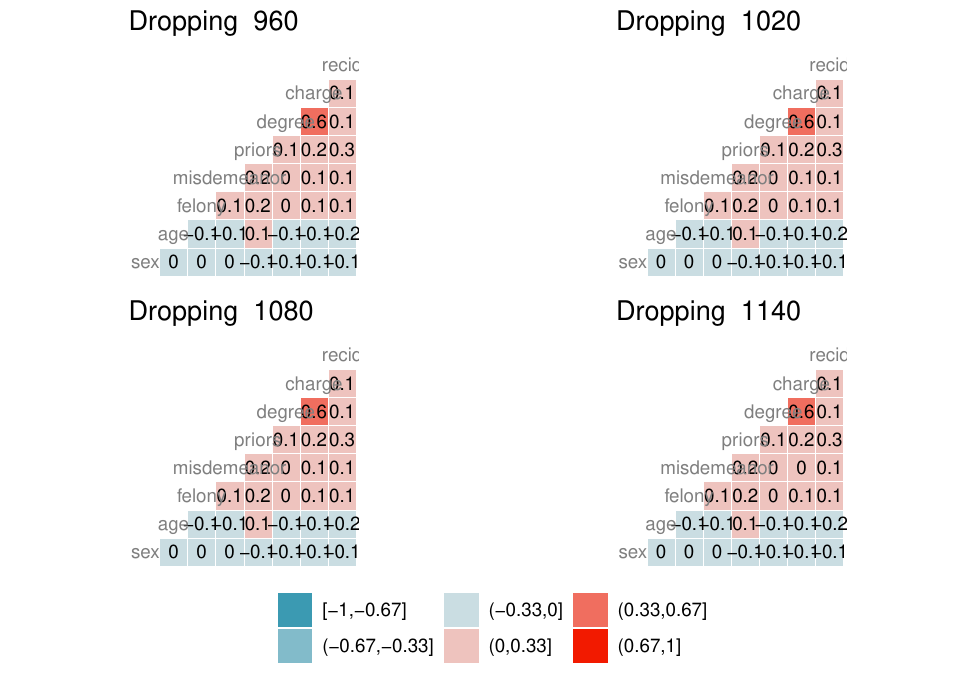}
\includegraphics[width=.5\textwidth]{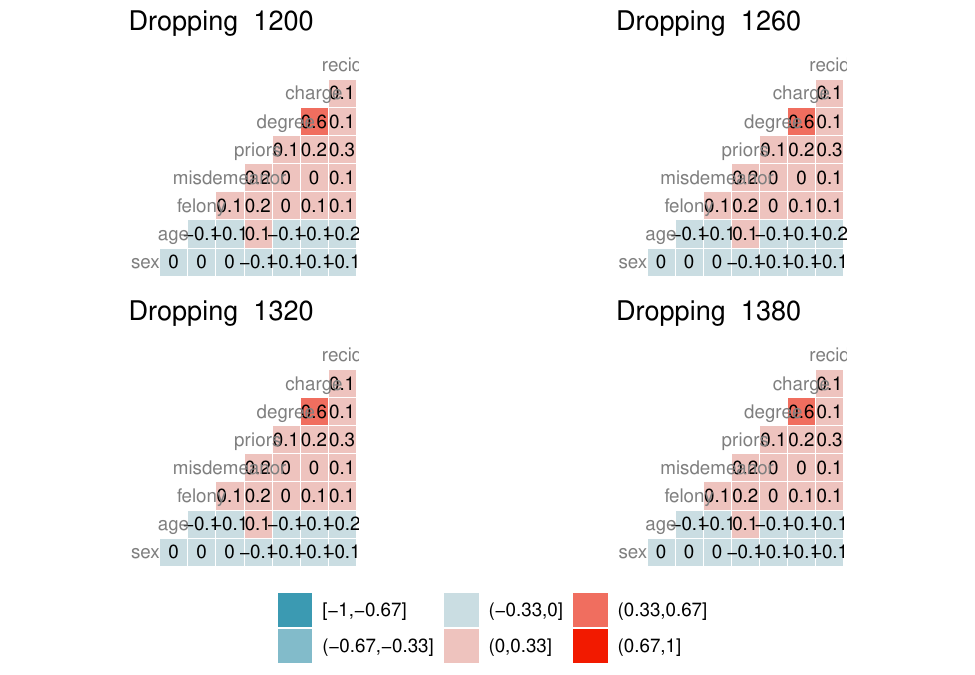}
\end{figure}

\begin{figure}
\caption{Correlation Heat Maps for Different Data Samples.}
\includegraphics[width=.5\textwidth]{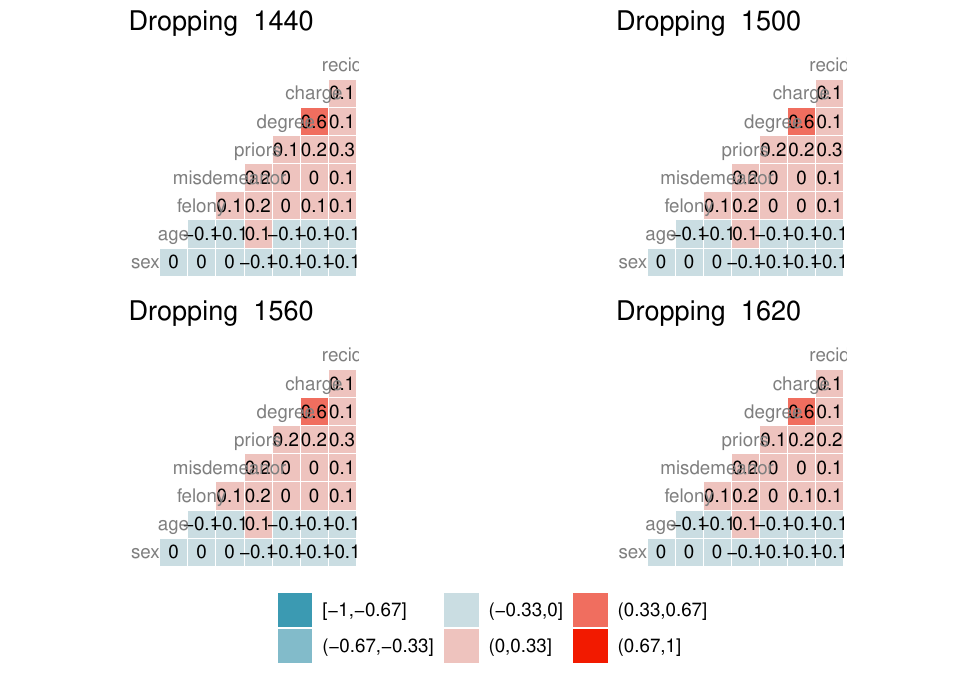}
\includegraphics[width=.5\textwidth]{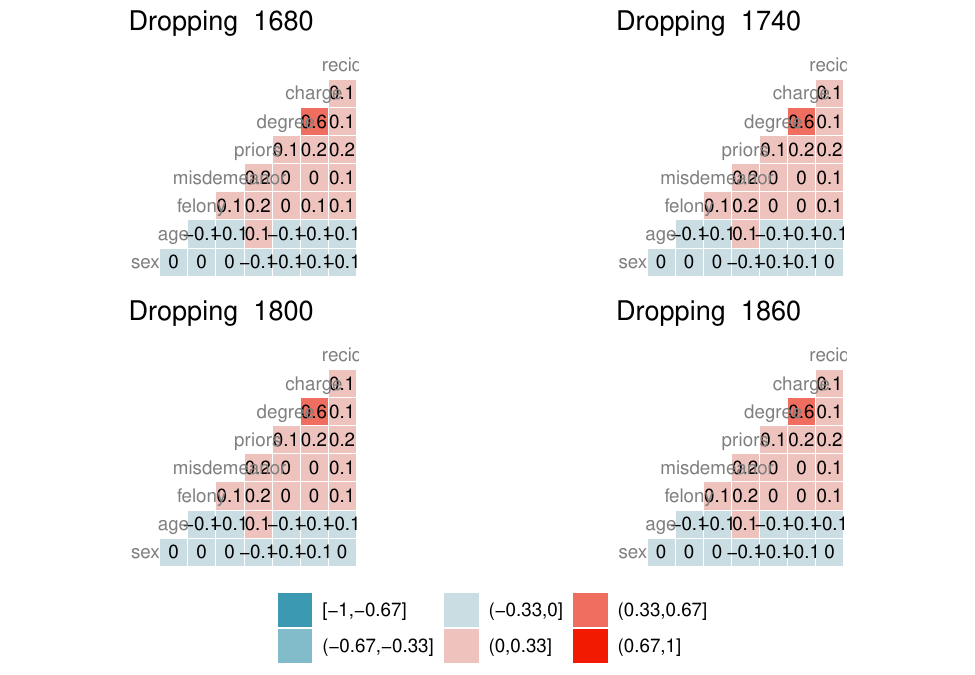}
\includegraphics[width=.5\textwidth]{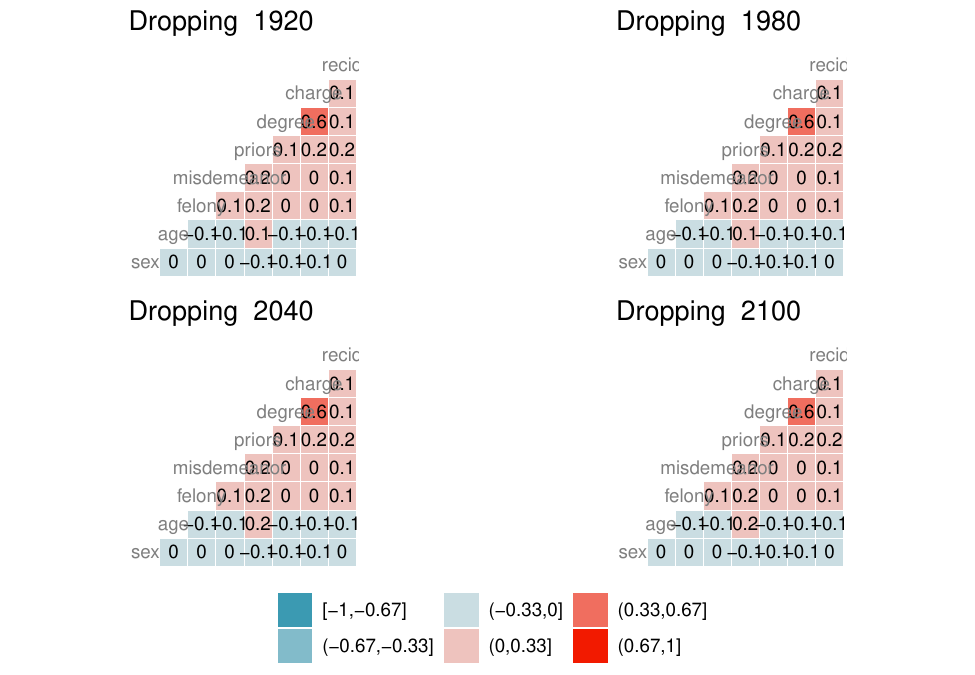}
\includegraphics[width=.5\textwidth]{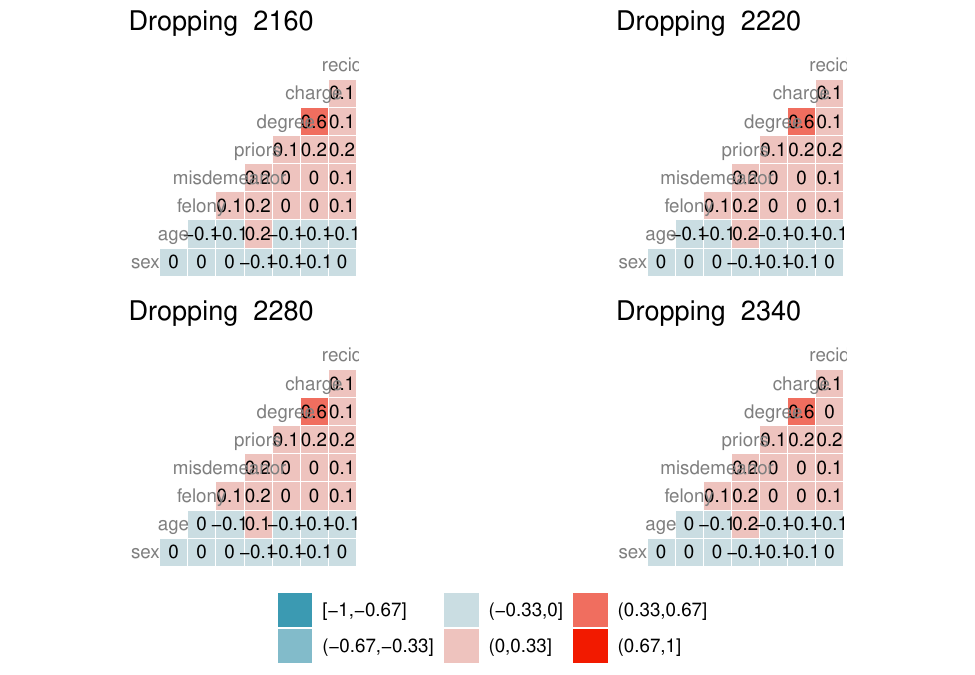}
\includegraphics[width=.5\textwidth]{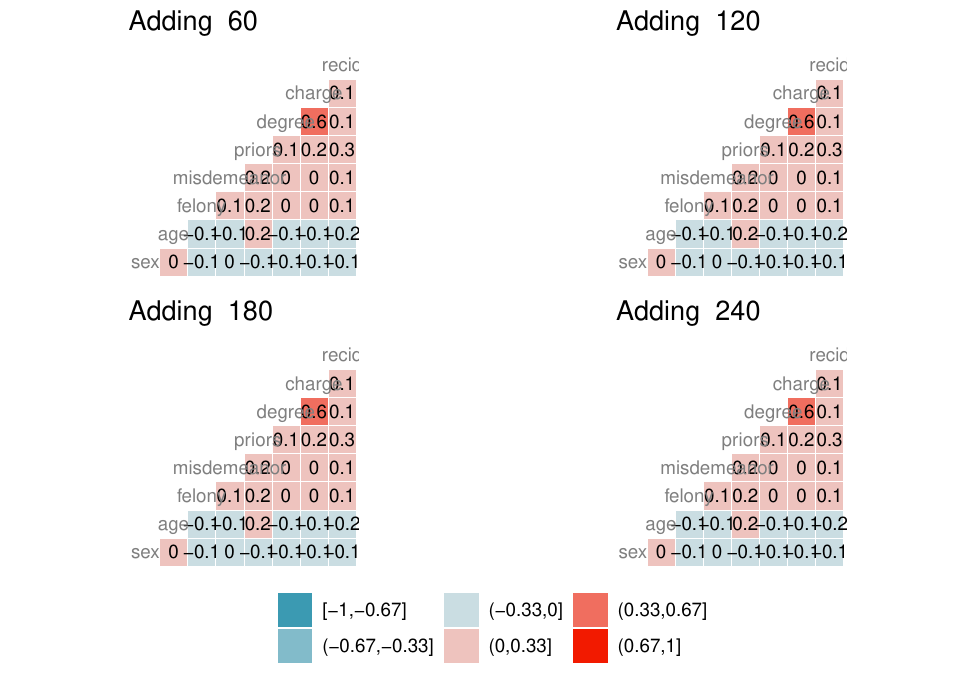}
\includegraphics[width=.5\textwidth]{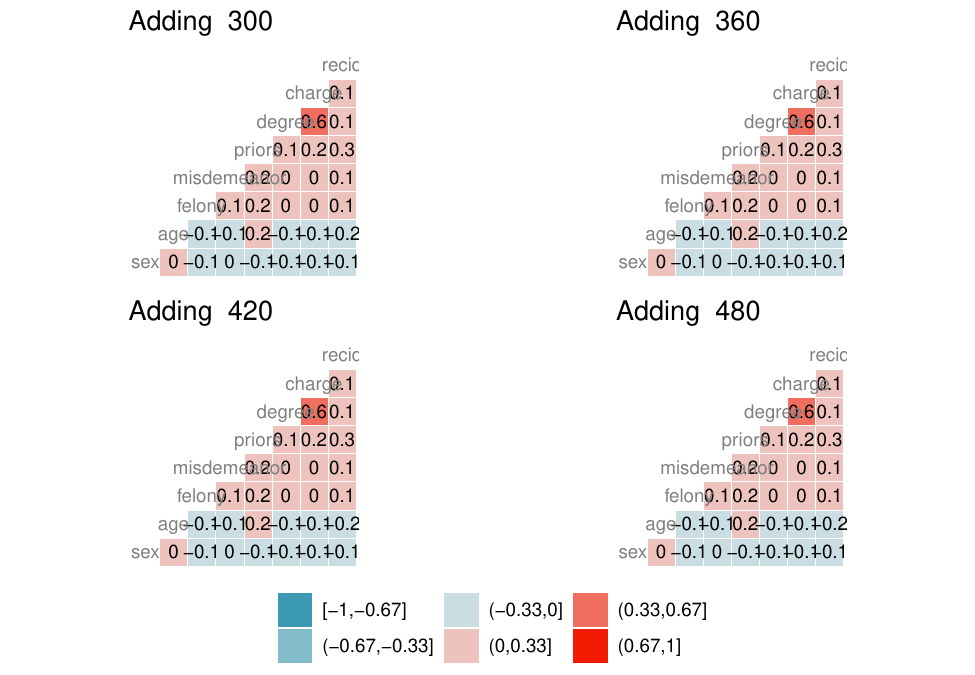}
\end{figure}

\begin{figure}
\caption{Correlation Heat Maps for Different Data Samples.}
\includegraphics[width=.5\textwidth]{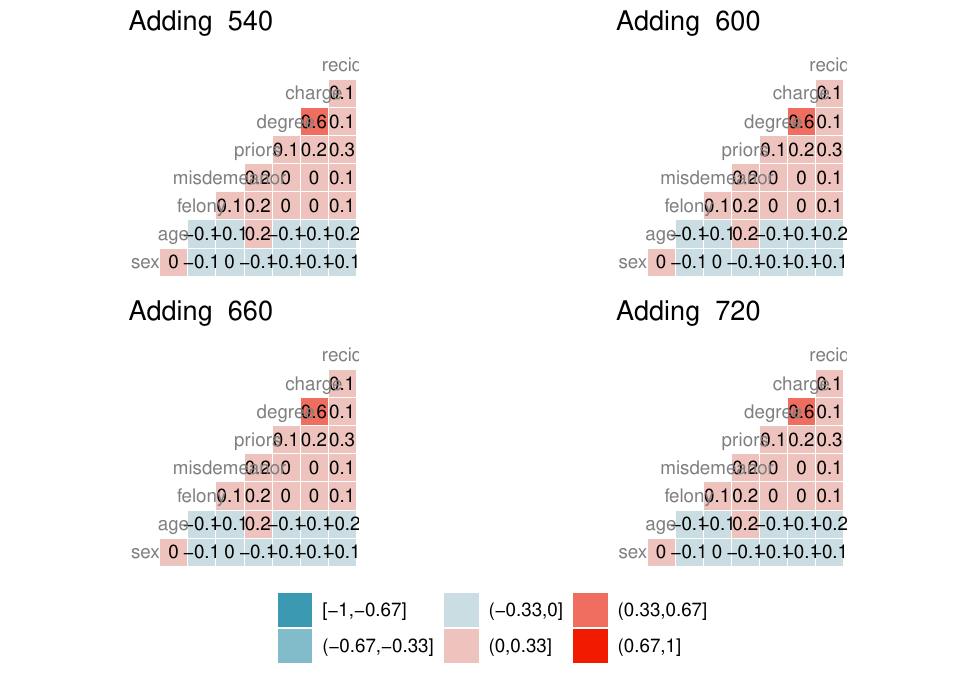}
\includegraphics[width=.5\textwidth]{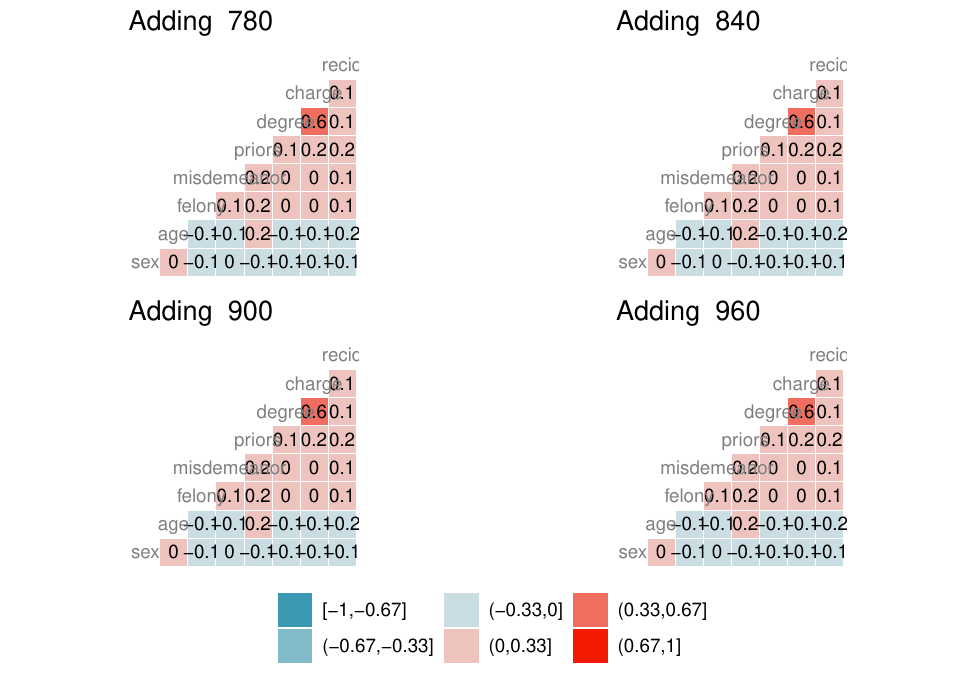}
\includegraphics[width=.5\textwidth]{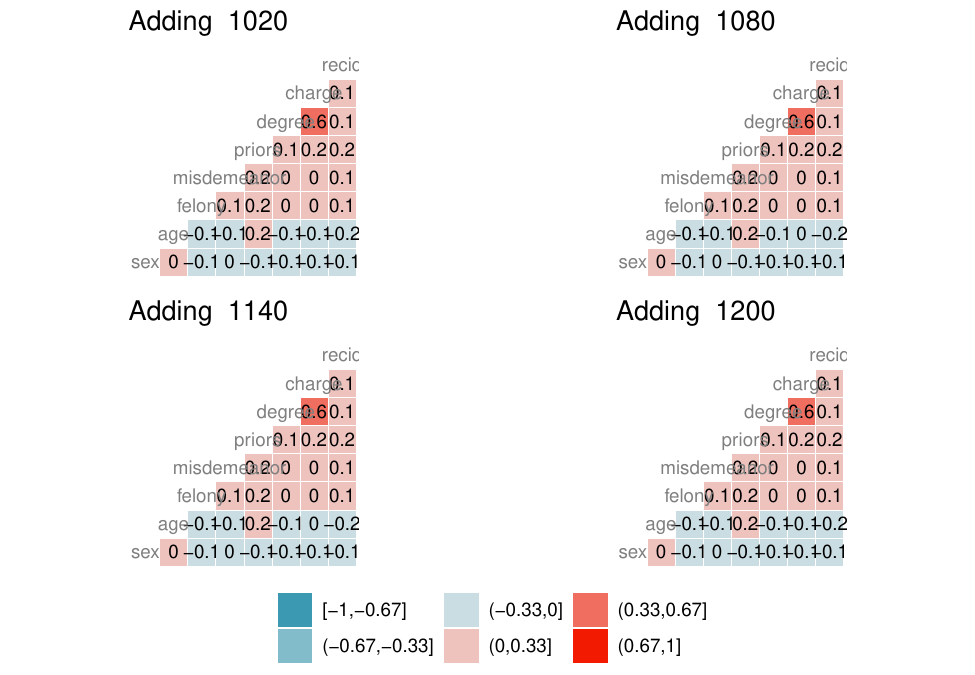}
\includegraphics[width=.5\textwidth]{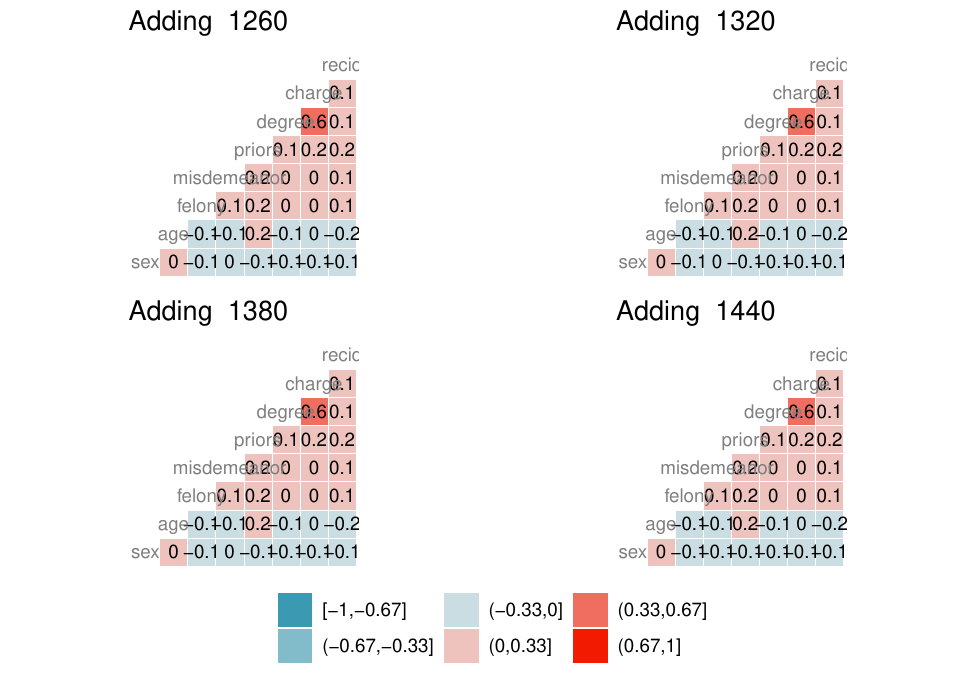}
\includegraphics[width=.5\textwidth]{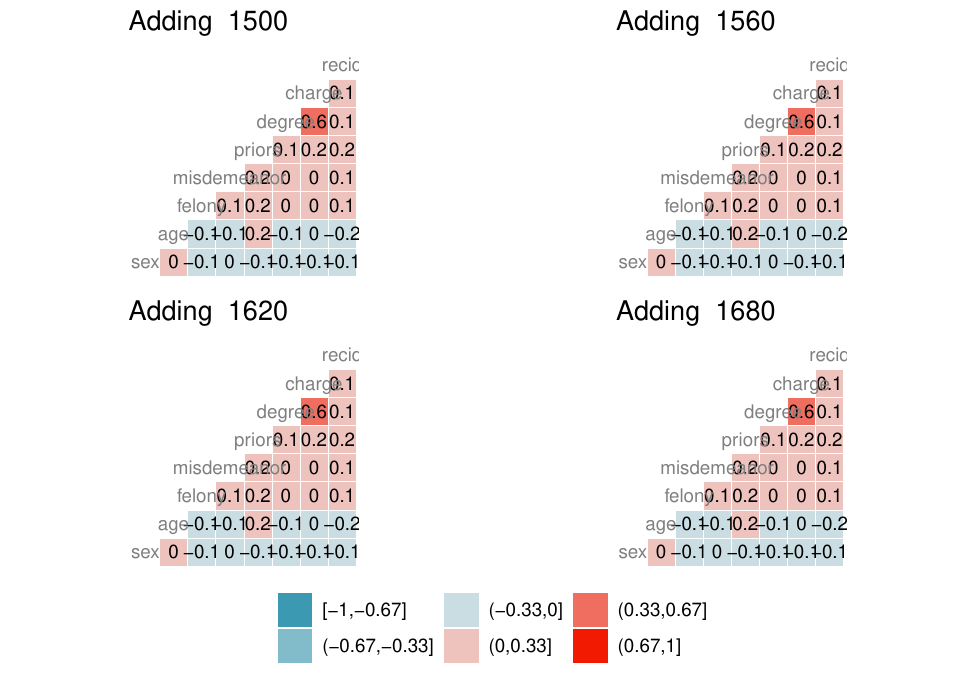}
\includegraphics[width=.5\textwidth]{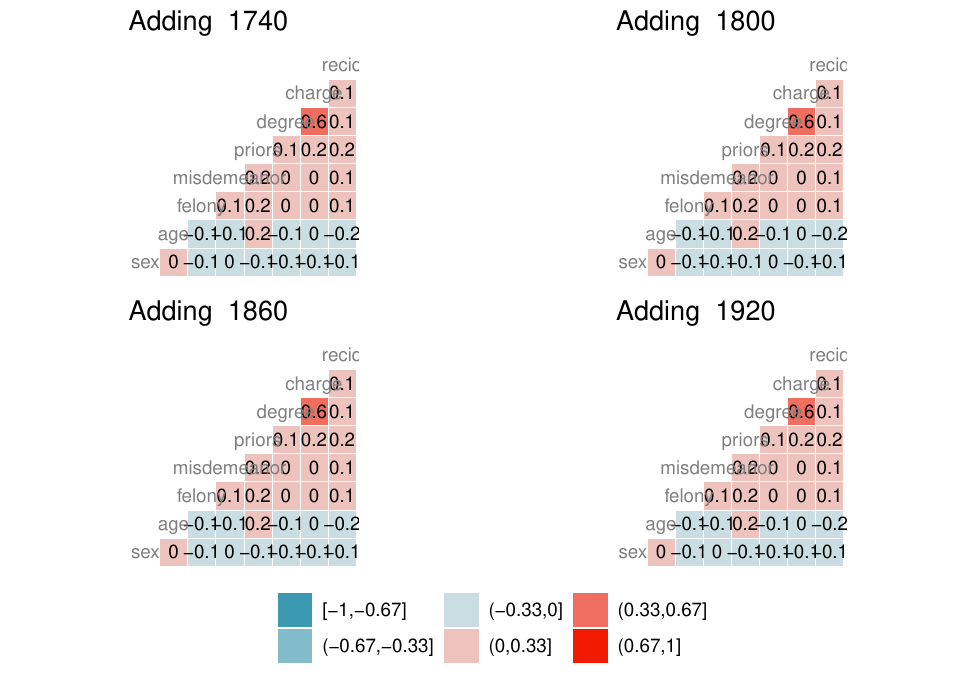}
\includegraphics[width=.5\textwidth]{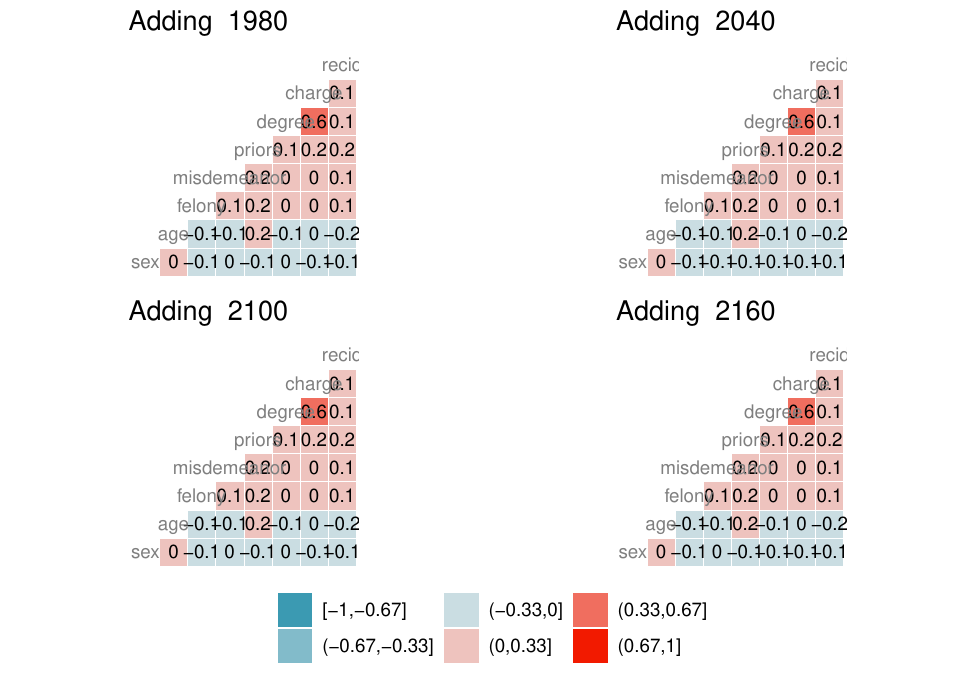}
\includegraphics[width=.5\textwidth]{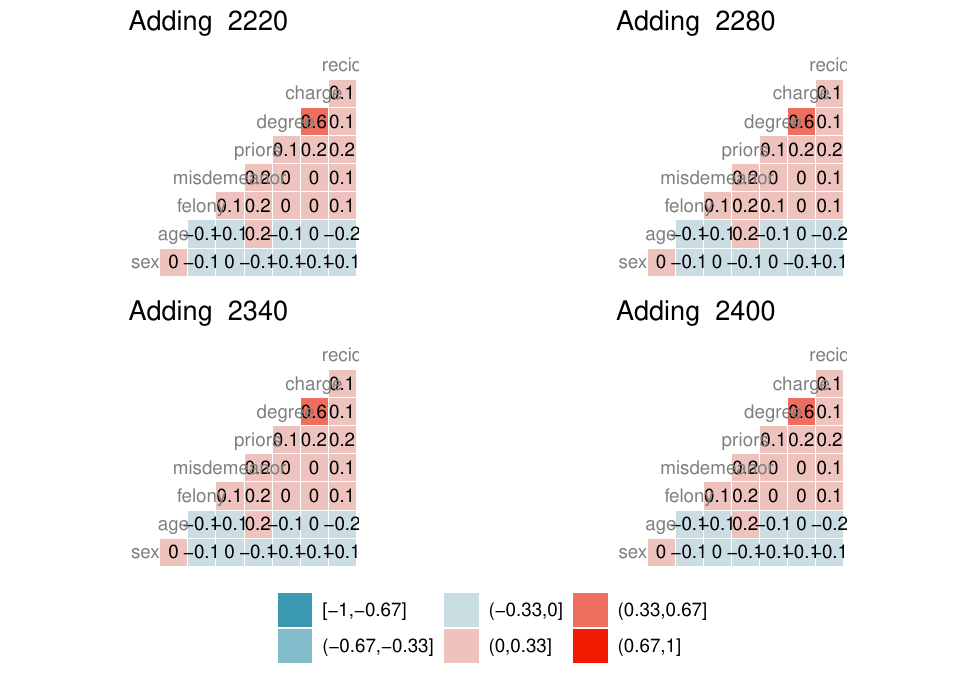}
\end{figure}

\end{document}